\documentclass{article}

\usepackage[numbers,sort&compress]{natbib}
\usepackage[preprint]{neurips_2025}

\usepackage{amssymb}
\usepackage{amsmath,amsfonts}
\usepackage{amsopn}
\usepackage{bm} %
\usepackage{multirow}
\usepackage{tabularx}
\usepackage{float} 
\newcommand{\vct}[1]{\boldsymbol{#1}} %
\newcommand{\mat}[1]{\boldsymbol{#1}} %

\newcommand{\OursA}{\method{SST}\xspace}
\newcommand{\OursL}{\method{OC-CCL}\xspace}
\newcommand{\field}[1]{\mathbb{#1}}

\newcommand{\R}{\field{R}} %
\newcommand{\eg}{\emph{e.g.}\xspace}
\newcommand{\ie}{\emph{i.e.}\xspace}

\newcommand{\ProbOpr}[1]{\mathbb{#1}}

\newcommand{\expect}[2]{%
\ifthenelse{\equal{#2}{}}{\ProbOpr{E}_{#1}}
{\ifthenelse{\equal{#1}{}}{\ProbOpr{E}\left[#2\right]}{\ProbOpr{E}_{#1}\left[#2\right]}}} %
\newcommand{\var}[2]{%
\ifthenelse{\equal{#2}{}}{\ProbOpr{VAR}_{#1}}
{\ifthenelse{\equal{#1}{}}{\ProbOpr{VAR}\left[#2\right]}{\ProbOpr{VAR}_{#1}\left[#2\right]}}} %

\newcommand{\vx}{{\vct{x}}}
\newcommand{\vy}{\vct{y}}

\newcommand{\mB}{\mat{B}}

\newcommand{\method}[1]{\textsc{#1}}
\newcommand{\eat}[1]{}

\usepackage[utf8]{inputenc} 
\usepackage[T1]{fontenc}    
\usepackage{hyperref}       
\usepackage{url}            
\usepackage{booktabs}       
\usepackage{amsfonts}       
\usepackage{nicefrac}       
\usepackage{microtype}      
\usepackage{xcolor}         
\usepackage{xspace}
\usepackage{graphicx}
\usepackage{enumitem}
\usepackage{caption}
\usepackage{wrapfig}
\usepackage{subcaption}
\usepackage[nameinlink,capitalize]{cleveref}

\title{\emph{Static Segmentation by Tracking}: \\A Label-Efficient Approach for Fine-Grained Specimen Image Segmentation} 

\author{
    \normalsize
    {Zhenyang Feng}\textsuperscript{1}, {Zihe Wang}\textsuperscript{1}, {Jianyang Gu}\textsuperscript{1},  {Saul Ibaven Bueno}\textsuperscript{1}, {Tomasz Frelek}\textsuperscript{1}, 
    \\ \normalsize
    {Advikaa Ramesh}\textsuperscript{1}, 
    {Jingyan Bai}\textsuperscript{1},
    {Lemeng Wang}\textsuperscript{1}, 
    {Zanming Huang}\textsuperscript{1}, {Jinsu Yoo}\textsuperscript{1},
    {Tai-Yu Pan}\textsuperscript{1}, 
    \\ \normalsize 
    {Arpita Chowdhury}\textsuperscript{1}, {Michelle Ramirez}\textsuperscript{1}, 
    {Elizabeth G. Campolongo}\textsuperscript{1},
    {Matthew J. Thompson}\textsuperscript{1}, {Christopher G. Lawrence}\textsuperscript{2}, 
    {Sydne Record}\textsuperscript{3}, {Neil Rosser}\textsuperscript{4}, 
    {Anuj Karpatne}\textsuperscript{5}, {Daniel Rubenstein}\textsuperscript{2}, {Hilmar Lapp}\textsuperscript{6}, {Charles V. Stewart}\textsuperscript{7},
    {Tanya Berger-Wolf}\textsuperscript{1},
    {Yu Su}\textsuperscript{1},
    {Wei-Lun Chao}\textsuperscript{1}
    \vspace{10pt}
    \\ \normalsize
    \textsuperscript{1}The Ohio State University, \textsuperscript{2}Princeton University, \textsuperscript{3}University of Maine, \textsuperscript{4}University of Miami, \textsuperscript{5}Virginia Tech,
    \textsuperscript{6}Duke University,
    \textsuperscript{7}Rensselaer Polytechnic Institute
    \vspace{10pt}
    \\ \small
    \url{https://github.com/Imageomics/SST}
}

\begin{document}
\maketitle
\begin{abstract}
We study image segmentation in the biological domain, particularly trait segmentation from specimen images (\eg, butterfly wing stripes, beetle elytra). This fine-grained task is crucial for understanding the biology of organisms, but it traditionally requires manually annotating segmentation masks for hundreds of images per species, making it highly labor-intensive.
To address this challenge, we propose a label-efficient approach, \textbf{Static Segmentation by Tracking (\OursA)}, based on a key insight: while specimens of the same species exhibit natural variation, the traits of interest show up consistently. This motivates us to concatenate specimen images into a  ``pseudo-video'' and reframe trait segmentation as a \textbf{tracking} problem. Specifically, \OursA generates masks for unlabeled images by propagating annotated or predicted masks from the ``pseudo-preceding'' images. Built upon recent video segmentation models, such as Segment Anything Model 2, \OursA achieves high-quality trait segmentation with only \textbf{one labeled image per species}, marking a breakthrough in specimen image analysis. To further enhance segmentation quality, we introduce a \textbf{cycle-consistent loss} for fine-tuning, again requiring only one labeled image. Additionally, we demonstrate the broader potential of \OursA, including one-shot instance segmentation in natural images and trait-based image retrieval.
\end{abstract} 
    
\section{Introduction}
\label{s:intro}
Understanding the sources and patterns of intra-specific variation in traits (\eg, morphological characteristics such as fin length in fish or wing size in beetles) is a central goal of evolutionary and ecological study~\cite{darwin1859origin, bolnick2011intraspecific}. Intra-specific trait variation provides a currency for assessing the roles of abiotic and biotic processes in community assembly, as it reflects the mechanisms driving species occurrence and their responses to change~\cite{violle2012return}. Museum specimens present an untapped resource for curating information on intra-specific trait variation in species morphology. Up until now, it has been difficult to harvest trait information from museum specimens due to the sheer amount of manual labor needed to make such measurements. Automatic segmentation of morphological traits from specimen images has the potential to scale up the measurement of traits and free up researchers to focus on analysis and interpretation. This paper originated from an interdisciplinary collaboration between biologists and computer scientists, aiming to segment images of organismal specimens to measure variation in traits to fill this much-needed knowledge gap.

\begin{figure*}[!ht]
\centering
\includegraphics[width=1\linewidth]{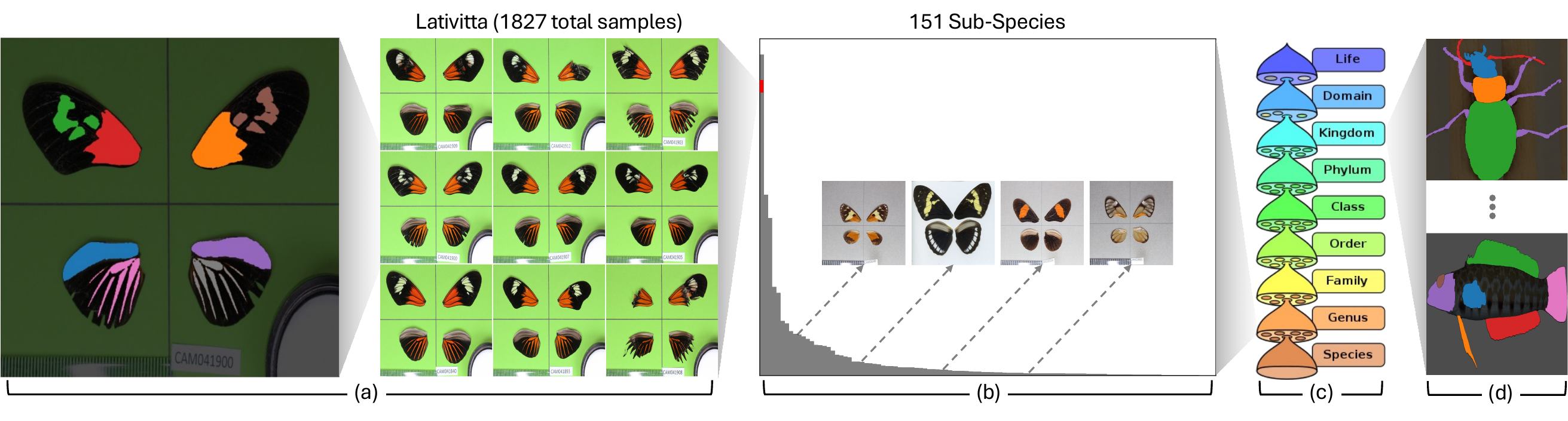}
\vskip-10pt
\caption{\small \textbf{Illustration of the trait segmentation problem from specimen images.} (a) Specimen samples of \textit{Heliconius erato lativitta}, and one example of segmentation masks. (b) The histogram of sample counts per \emph{subspecies} in the Cambridge Butterfly Collection~\cite{lawrence2024heliconius}, with exemplar images. (c) These subspecies belong to the genus \textit{Heliconius}, which falls under the suborder  \textit{Rhopalocera}, encompassing over $10,000$ butterfly species worldwide.  
(d) Trait segmentation is also important for other animals, such as beetles and fish.}
\vskip-4pt
\label{fig-2: specimen}
\end{figure*}

\begin{figure*}[!ht]
\centering
\includegraphics[width=1\linewidth]{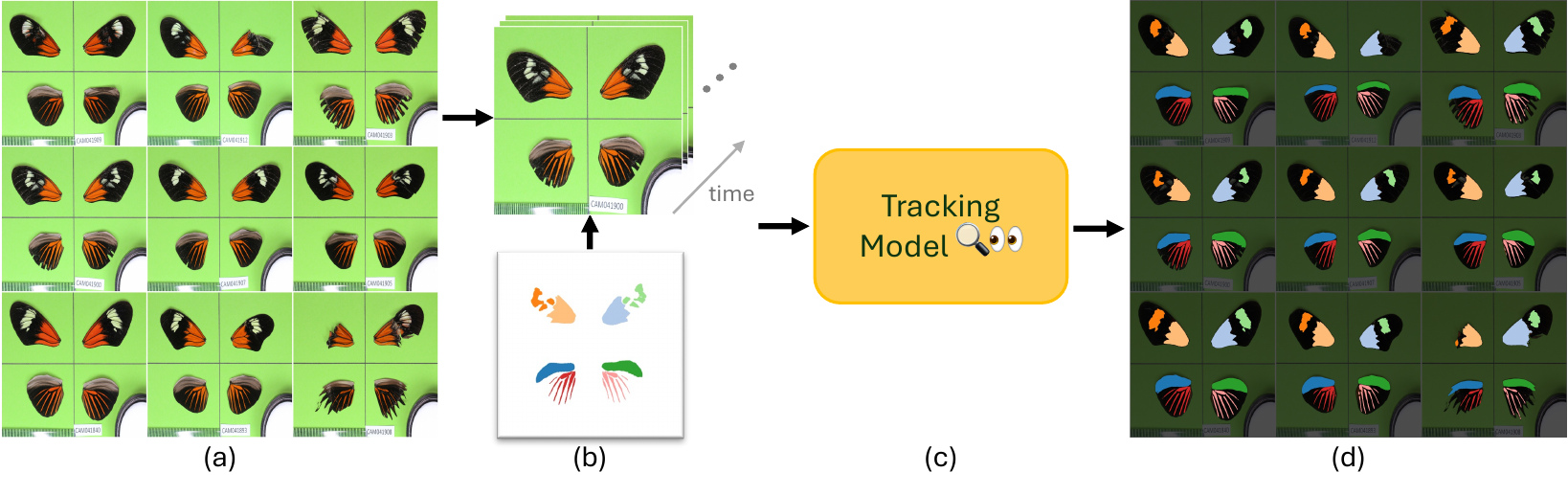}
\vskip-12pt
\caption{\small \textbf{Illustration of our approach Static Segmentation by Tracking (\OursA).} (a) Different specimens of the same species. (b) We concatenate these static, non-sequential images into a pseudo-video. (c) The annotated masks of the first image are treated as the prompt to a tracking algorithm, such as SAM~2~\cite{ravi2024sam}. (d) \OursA can achieve high-quality trait segmentation in a one-shot manner.}
\vskip-14pt
\label{fig-3: SST}
\end{figure*}

Training a segmentation model \cite{minaee2021image,li2024transformer,liu2021review, feng2020deep,chandrakar2022animal} is arguably the most straightforward approach to this problem. 
However, it requires annotating traits on tens, if not hundreds, of images per species to ensure the model generalizes well. This process is itself laborious, let alone there are millions of species on Earth and many of them do not have sufficient samples for labeling (see~\cref{fig-2: specimen}). Several recent segmentation algorithms focused on a few-shot setting, aiming to adapt models to new concepts using only a handful of labeled examples~\cite{wang2019panet,hong2021cost,wang2023seggpt,tian2020pfenet}. However, most of these methods were designed to segment a single concept at a time (\eg, an entire beetle) rather than multiple traits jointly (\eg, the beetle's head, antennae, and elytra). Even when segmenting a single trait, they often struggle to capture fine details, performing much worse than many-shot methods (see \cref{s:exp}). We thus ask, 
\begin{center}
\vskip -4pt
\color{blue}
\emph{
    How can we perform fine-grained segmentation on specimen images\\ without a large amount of labeled data? 
}
\end{center}

We begin with a deeper look at specimen images, particularly those of the same species. We make several key observations (see \cref{fig-2: specimen}). From a \emph{macro} 
perspective, where a specimen is viewed as a ``whole,'' biological variations naturally cause specimens of the same species to appear different; some may even have damaged parts.
However, from a \emph{micro} perspective, where a specimen is seen as a ``composition'' of traits---\emph{the components we aim to segment}---specimens of the same species look quite similar in their trait layouts. Unless damaged, these traits consistently appear and maintain structured spatial relationships with one another. Notably, each trait has distinct characteristics, such as color, shape, size, pattern, and relative position, offering rich cues for identifying and locating them across specimens of the same species.

Building on these insights, we propose \textbf{reframing trait segmentation in specimen images as a tracking problem}. Tracking involves identifying an initial set of instances, assigning each a unique ID, and following them across video frames~\cite{luo2021multiple,yilmaz2006object}. \emph{In our case, the instances are distinct traits, each marked by a unique color as in \cref{fig-3: SST} (b).} 
While we do not have a true video, but rather a set of static specimen images, the variations observed in traits across images---such as changes in size, location, orientation, shape, and color---closely resemble the transformations seen in video frames due to camera movement, motion, deformation, and lighting changes. Even damaged parts can be viewed as occlusions in this analogy.
This motivates us to concatenate  \emph{static, non-sequential} specimen images into a ``pseudo-video'' and apply a tracking algorithm to locate and segment individual traits across frames---\textbf{given only the annotated segmentation masks from the first frame} (\cref{fig-3: SST}). 
 
We name our approach \textbf{Static Segmentation by Tracking (\OursA)}, which \emph{lifts an image segmentation problem into a tracking problem, leveraging the latter's characteristics to achieve the former in a remarkably label-efficient, one-shot manner}. 
Essentially, the model's task is to localize each annotated mask from the first frame in subsequent frames and then \emph{propagate} and \emph{deform} it accordingly.

\begin{figure}[t]
  \centering
  \begin{minipage}[c]{0.4\textwidth}
  \includegraphics[width=\linewidth]{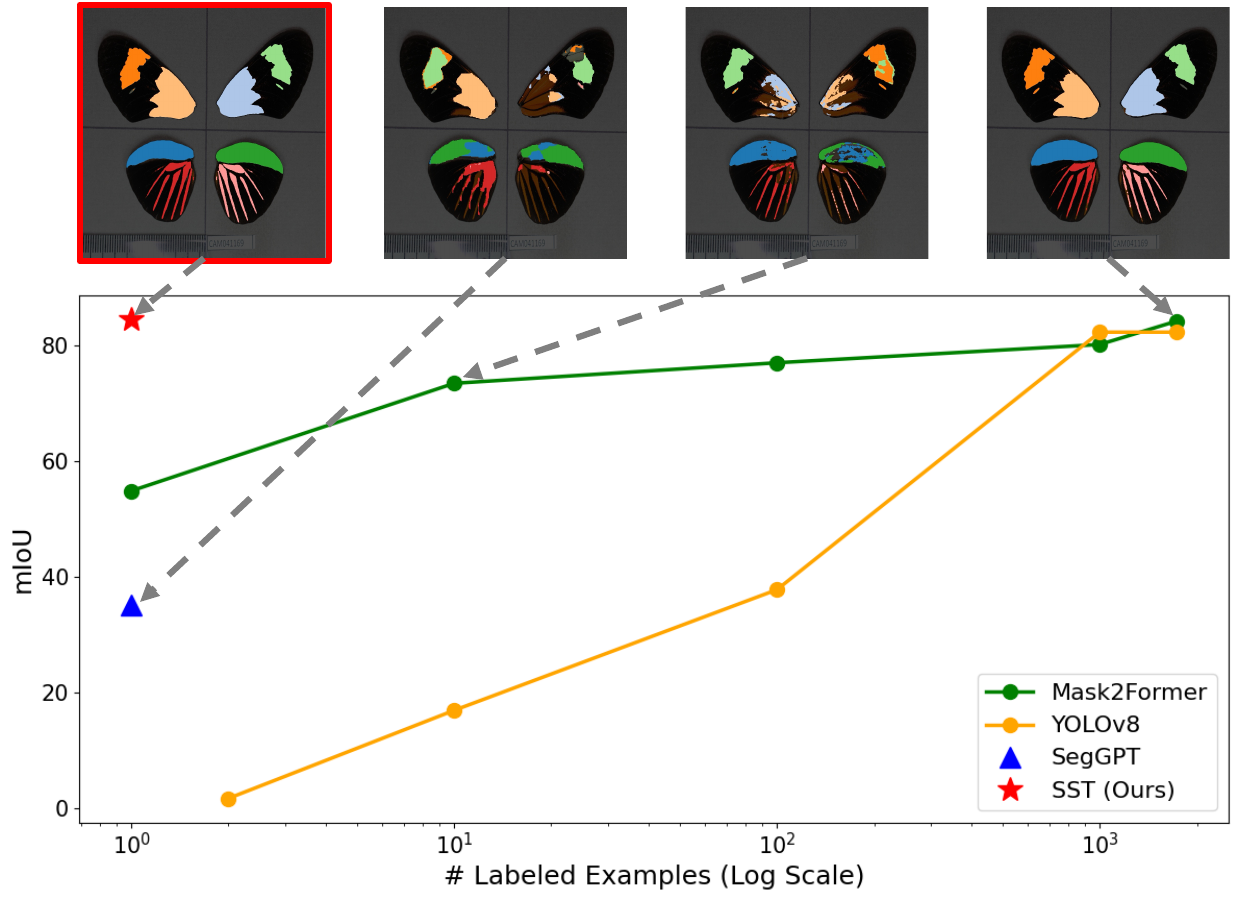}
  \caption{\small \textbf{Static Segmentation by Tracking (\OursA)} outperforms other one/many-shot baselines (on \textit{Heliconius erato lativitta}).}
  \label{fig-1: performance comparison}
  \end{minipage}
  \hskip 10pt
  \begin{minipage}[c]{0.55\textwidth}
  \includegraphics[width=\linewidth]{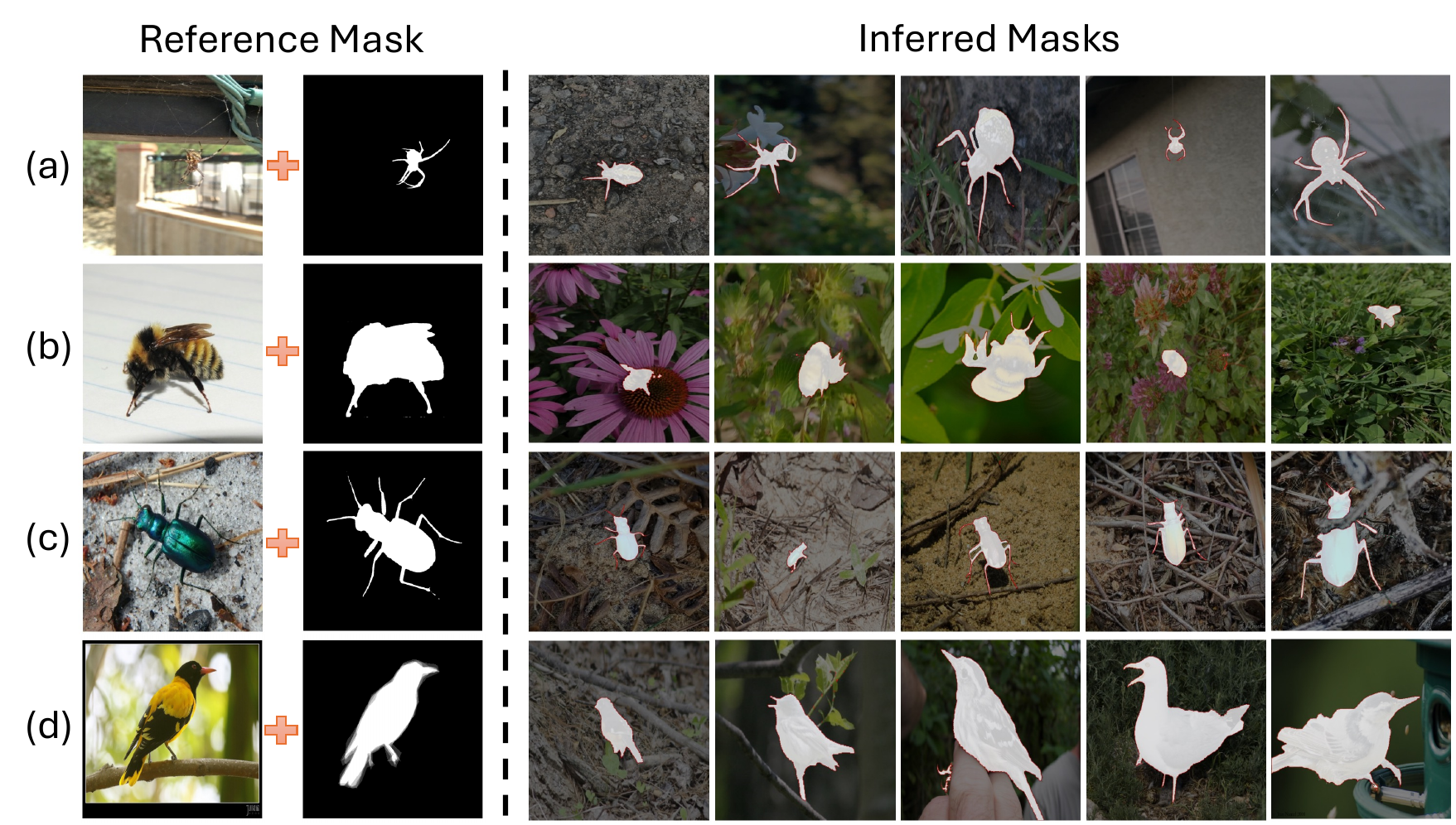}
  \caption{\small \textbf{\OursA applies to in the wild images.} (a–c) Spiders, bees, and beetles in iNaturalist~\cite{inaturalist-2021}. (d) Birds in CUB~\cite{wah2011caltech}.}
  \label{fig-4: bird and spider}
  \end{minipage}
  \vspace{-15pt}
\end{figure}

We implement \OursA using recent pre-trained video segmentation models, including Cutie~\cite{cheng2024putting}, DEVA~\cite{cheng2023tracking}, and SAM~2~\cite{ravi2024sam}. Given the annotated masks from the first frame as prompts, these models are capable of tracking them across frames.
We evaluate \OursA on three specimen image datasets: Cambridge Butterfly~\cite{lawrence2024heliconius}, NEON Beetle~\cite{Fluck2018_NEON_Beetle}, and Fish-Vista \cite{fishvistaData}. \OursA demonstrates much better performance than other one-shot baselines, such as SegGPT~\cite{wang2023seggpt}, in trait segmentation.  Surprisingly, in some scenarios, \OursA even outperforms segmentation models trained with abundant labeled data, including Mask2Former~\cite{cheng2022masked} and YOLOv8~\cite{Jocher_Jing_Chaurasia} (see~\cref{fig-1: performance comparison}). \emph{We attribute this success to the fact that \OursA does not treat labeled and unlabeled images as IID samples---an assumption underlying most image segmentation algorithms---but instead explicitly leverages their dependency to facilitate segmentation.} We view this as a breakthrough in the analysis of specimen images.

\textbf{Further improvement (\cref{3.2. OC-CCl}).} \OursA uses pre-trained models, and as such, it may struggle when transitions between static specimen images differ significantly from those in the training videos. To address this, we propose the \textbf{Opening-Closing Cycle-Consistent Loss (\OursL)} for \emph{semi-supervised model fine-tuning}, leveraging the same labeled image as the prompt for supervision.

\textbf{Further exploration (\cref{ss:further_explore}).} We explore additional application scenarios for \OursA. Beyond specimen images, we also find that \OursA performs well on \textbf{instance segmentation of animals} in natural images, even when the ``pseudo-video'' contains rapidly changing and arbitrarily varying backgrounds (\cref{fig-4: bird and spider}). Additionally, we investigate \textbf{trait-based image retrieval}, aiming to retrieve specimen images that share a specific trait with the query image. By repurposing \OursL, we demonstrate that \OursA can accurately retrieve images containing specific traits, such as the white bands on butterfly forewings (\cref{fig-10: trait retrieval}).

\textbf{Contribution.} 
In addition to \OursA and \OursL, we present following contributions:
We hand-labeled over $813$ butterfly specimen images and semi-automatically labeled $2,831$ images with trait masks, covering more than $150$ subspecies. We also hand-labeled $180$ beetle specimen images. These labeled datasets are intended to serve as a testbed for future research in specimen image segmentation.
We demonstrate the potential of \OursA in broader application scenarios, including instance segmentation of images taken in the wild and trait-based image retrieval.

\textbf{Remark.} 
This paper originated from an interdisciplinary collaboration aimed at addressing a real-world bottleneck in specimen image analysis. The primary challenge arises from the fine-grained nature of traits, the labor-intensive annotation process, and the vast diversity of species.
Since species-specific methods are not scalable, we focus on identifying common properties that enable a more generalizable solution, ideally leveraging recent foundation models.

Our proposed approach, \OursA, embodies this principle by effectively leveraging dependencies across samples, resulting in a solution that is both simple and generalizable to tackle this long-standing challenge. While~\OursA might appear straightforward \emph{in hindsight}, its development was far from trivial. Conventional image segmentation models remain the dominant approach, yet they struggle with scalability and adaptability across species. The key novelty of our work lies in recognizing and implementing the \emph{appropriate} way to address specimen image segmentation---one that transcends the label-intensive, species-specific constraints in favor of a more generalizable framework. 
Please also refer to \cref{sec: conclusion} for a discussion on the paper's scope.

\section{Related Work and Background}
\label{s:related}

\noindent\textbf{Image segmentation} is a long-standing challenge in computer vision~\cite{liu2021review, feng2020deep,chandrakar2022animal,minaee2021image,li2024transformer}. Semantic and instance segmentation are among the most popular tasks today, aiming to group pixels with the same semantic meanings and instances \cite{minaee2021image, hafiz2020survey}. While much of the focus has been on segmenting common, coarse-grained objects, recent works have begun exploring part-level segmentation within these objects \cite{Hung_2019_CVPR, Pan_2023_CVPR}. In this paper, we address an underexplored challenge: trait segmentation for fine-grained object categories, such as subspecies of butterflies.

While state-of-the-art (SOTA) models have shown impressive capabilities in segmentation \cite{cheng2022masked, Jocher_Jing_Chaurasia}, collecting pixel-level annotations for training is labor-intensive. To address this, \textbf{few-shot segmentation (FSS)} has emerged as a promising paradigm, attempting to use few-shot learning techniques to generate high-quality segmentation masks for new classes \cite{tian2020pfenet,liu2020crnet,zhang2019canet,xue2024hdmnet,hong2021VAT}. Here, we propose a novel perspective and algorithm to tackle the FSS problem. 

\textbf{Video segmentation} focuses on segmenting the same concepts (\eg, objects) across video frames~\cite{yao2020video}. Compared to image segmentation, it requires associating masks between frames to assign the same labels. Recent models~\cite{ravi2024sam,cheng2024putting,cheng2023tracking} are mainly built upon the transformer architecture with memories, and trained on a large-scale video data. Given annotated masks from the first frame, they can track and segment the target instances in subsequent frames. 

While image and video segmentation have typically been studied separately, we show that models developed for the latter can be effectively applied to the former in a label-efficient manner, even when the images are non-sequential.

\textbf{Co-segmentation and image registration} share similar properties with our approach---they leverage dependencies across images. Co-segmentation locates objects that appear in multiple images in an unsupervised manner~\cite{rother2006cosegmentation,9057736,joulin2010discriminative,Cho_2015_CVPR, vo2020toward}. Our work can be viewed as a one-shot supervised approach, leveraging tracking models to efficiently segment the traits of interest across images. Image registration is widely used to densely align pixels between images, such as brain MRI scans~\cite{mendrik2015mrbrains}. In our application, we do not need exact pixel-to-pixel associations; we only need to localize the traits of interest across images. 

\section{Proposed Approach}
\label{s:approach}

\textbf{Problem definition and notation.} We study trait segmentation from specimen images of the \emph{same} species.
Let $\vx\in\R^{W\times H \times 3}$ denote a $W\times H$ image and $\vy\in\{0, 1\}^{W\times H \times C}$ denote the corresponding ground-truth segmentation masks of $C$ distinct traits. The goal is to develop a segmentation model $f$ such that its output $\hat{\vy} = f(\vx)$ matches $\vy$.

Typically, one needs to collect a labeled training set with ample pairs of $(\vx, \vy)$, and use it to train $f$ in a supervised way. In this paper, we target the one-shot scenario, \ie, building $f$ using a single labeled image $(\vx, \vy)$.

\subsection{Static Segmentation by Tracking \textbf{(\OursA)}}
\label{ss:SST_basic}

At first glance, this seems like a daunting challenge. However, the domain-specific properties described in~\cref{s:intro} provide a crucial foundation. Our proposed approach, \textbf{\OursA}, leverages these properties by reframing trait segmentation as a tracking problem, which naturally becomes a one-shot task given a set of labeled instances in the first frame.

More specifically, let $\{\vx_0, \cdots, \vx_N\}$ denote a \emph{sequence} of video frames; a tracking algorithm aims to track the instances in $\vx_0$, encoded by the label $\vy_0$, across the remaining frames. In our context, we do not have a real video but rather a \emph{set} of $N$ unlabeled images and one labeled image $(\vx, \vy)$. Nevertheless, the domain-specific properties motivate us to construct a ``pseudo-video'' by treating $\vx$ as the first frame $\vx_0$, followed by the unlabeled images.

\textbf{\OursA with pre-trained models.} We leverage recent pre-trained video segmentation models for tracking~\cite{cheng2024putting,cheng2023tracking,ravi2024sam}. Despite differences in technical details, these models share a similar architecture. \emph{Without loss of generality, we focus on the Segment Anything Model 2 (SAM~2)} \cite{ravi2024sam} \emph{for the remainder of this section.} Below, we first briefly introduce its model architecture and inference mechanism.

\begin{wrapfigure}{r}{0.2\textwidth}
  \centering
  \includegraphics[width=\linewidth]{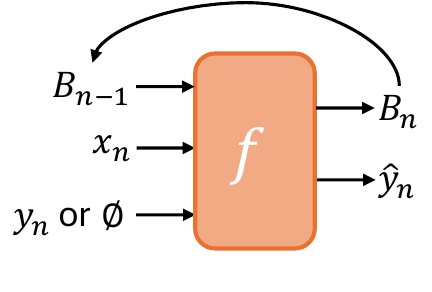}
  \vspace{-12pt}
  \caption{\small \textbf{SAM~2 inference mechanism.}}
  \label{fig-5: SAM2}
  \vspace{-10pt}
\end{wrapfigure}

SAM~2 uses a promptable Transformer encoder-decoder $f$ augmented with a memory bank $\mB$ to process a video and generate masks (see \cref{fig-5: SAM2}). 
Let $\{\vy_0, \cdots, \vy_N\}$ denote the ground-truth labels for video frames $\{\vx_0, \cdots, \vx_N\}$. When the label of $\vx_n$ is unavailable, we set $\vy_n=\varnothing$. Let  $\hat{\vy}_n$ denote the predicted mask for $\vx_n$, and let $\mB_n$ represent the updated memory bank after the prediction, which stores both the feature and mask information. $\mB_n$ can then be accessed by the next frame $\vx_{n+1}$ to connect consecutive frames. In the context of tracking, $\mB_n$ can be interpreted as the updated state estimate after perceiving the measurement $\vx_n$.

At each timestamp $n$, $f$ takes the tuple $[\vx_n, \mB_{n-1}, \vy_n]$ as input, where $\vy_n$ is treated as the (optional) prompt. It then outputs the tuple $[\mB_{n}, \hat{\vy}_n]$, where $\mB_{n}$ will be used as input at the next timestamp,
\[
[\mB_{n}, \hat{\vy}_n] = f\left([\vx_n, \mB_{n-1}, \vy_n]\right).    
\]

In our context, we have $\vy_n = \varnothing, \forall n > 0$, meaning only the first frame $\vx_0$ is annotated. By inputting $\vy_0$ to $f$ at timestamp $0$, we instruct the model on what to \emph{segment}---the distinct traits and their extents. The resulting memory bank $\mB_0$ then carries this information to successive frames, guiding the model on what to \emph{track} across frames to generate the masks $\{\hat{\vy}_1, \cdots, \hat{\vy}_N\}$. See~\cref{fig-3: SST} for an illustration.

\textbf{Pseudo-video creation.} There are multiple ways to concatenate unlabeled images into a pseudo-video. Intuitively, an order with smooth transitions is preferred, as it could potentially improve \OursA's performance. In contrast, non-smooth transitions may degrade \OursA. However, searching for an optimal order incurs additional computational cost.

To eliminate the uncertainty in creating pseudo-videos, we implement \OursA by constructing multiple short, two-frame videos, unless stated otherwise. Specifically, given the labeled image-mask pair $(\vx_0, \vy_0)$ and $N$ unlabeled images $\{\vx_1, \cdots, \vx_N\}$, we construct $\{\vx_0, \vx_1\}, \cdots, \{\vx_0, \vx_N\}$ and apply \OursA independently to each. 

We note that the above implementation also ensures a fair comparison to the baselines. Conventional image segmentation models process each test sample independently, reflecting the real-world online use case where a newly captured image is processed immediately. Recent few-shot approaches similarly process each test image independently. Noting that considering all test images jointly would transform the conventional \emph{inductive} setting into a \emph{transductive} one, we choose to process each test image independently.

That said, in the Appendix, we explore concatenating multiple unlabeled images. In the long run, developing an approach to search for the optimal order would be valuable.

\textbf{Remark.} According to the original paper~\cite{ravi2024sam}, SAM~2 is readily applicable to a batch of static, non-sequential images, \emph{by setting the memory bank $\mB_n$ to empty}. In essence, without the memory bank, SAM~2 treats each input image as an IID sample and processes them independently. 

Our insight is that even if the input images are taken in a non-sequential manner, whenever there exists a useful dependency among them (\eg, from the same species), SAM~2 has the potential to leverage this dependency. The key is to allow the memory bank to update, rather than resetting it.

\subsection{One-Shot Fine-Tuning for \textbf{\OursA}}
\label{3.2. OC-CCl}

\OursA uses pre-trained video models in a plug-and-play fashion without altering their weights, even though our use case might be outside the training data distribution. As a result, \OursA is expected to fail when transitions between static images are significantly out-of-distribution (OOD). 

One intuitive way to address this is model fine-tuning. However, with only one labeled image $(\vx_0, \vy_0)$, fine-tuning risks overfitting. Additionally, since we have used $\vy_0$ to prompt the model, we face another challenge: \emph{it is unclear how to use it ``dually'' as the label to supervise fine-tuning.}

\textbf{Opening-Closing Cycle-Consistent Loss (\OursL).}
To overcome these challenges, we propose a novel fine-tuning approach that leverages the flexibility of creating pseudo-videos. 
We can duplicate static images and inject them into the video sequence at different timestamps, allowing us to obtain multiple predictions for the same image.  
Specifically, given a short pseudo-video $\{\vx_0, \vx_1\}$, we duplicate both images, denoted by  $\dagger$, and create a palindrome-style cycle $\{\vx_0, \vx_1, \vx_1^\dagger, \vx_0^\dagger\}$, inspired by~\cite{jabri2020space}. 
In this cycle, the labeled image $\vx_0$ serves as both the ``opening'' and ``closing'' frames. Note that we do not require the label of $\vx_1$.

\begin{wrapfigure}{r}{0.45\textwidth}
  \vspace{-25pt}  
  \centering
  \includegraphics[width=\linewidth]{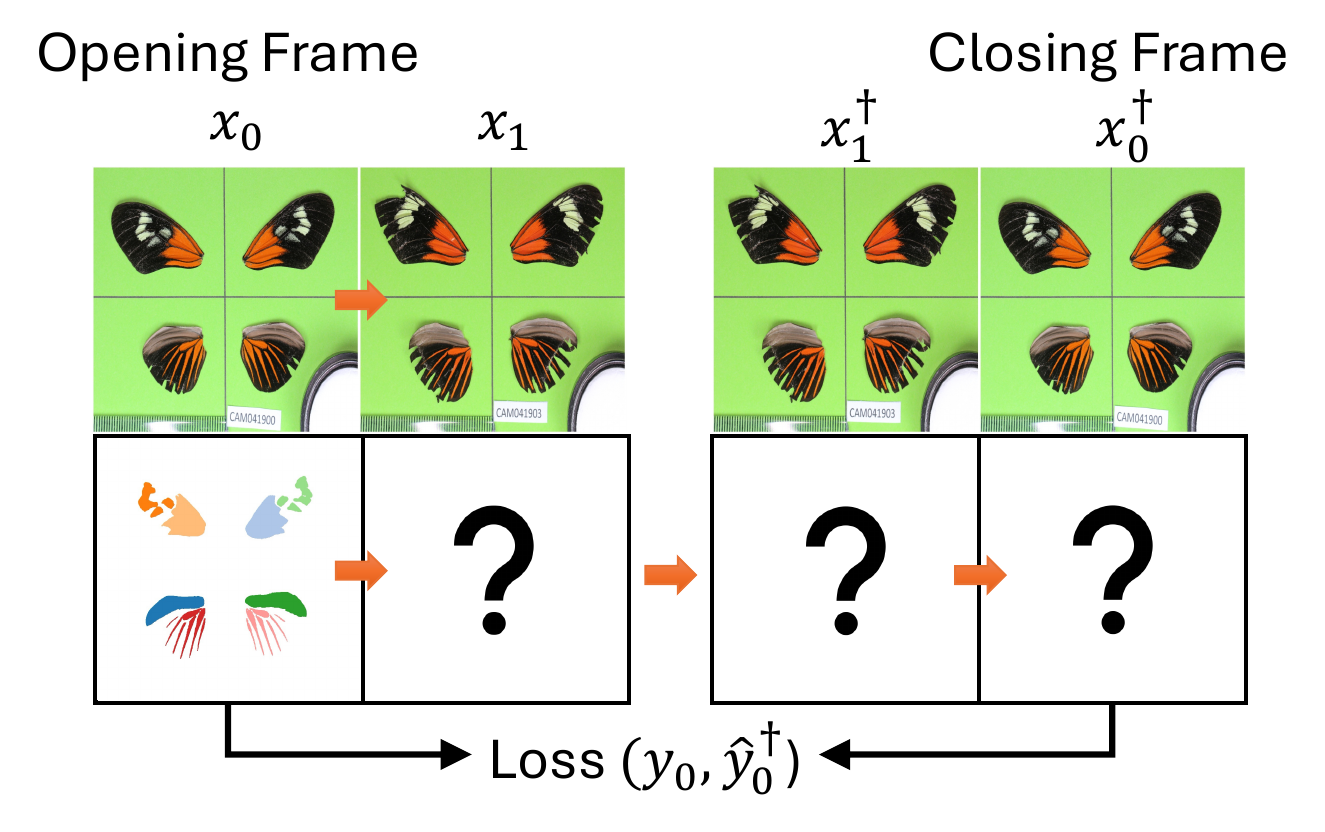}
  \vspace{-15pt}  
  \caption{\small \textbf{Opening-Closing Cycle-Consistent Loss (\OursL)} compares predicted masks to the ground truth.}
  \label{fig-6: CCL}
  \vspace{-25pt} 
\end{wrapfigure}

Unlike timestamp $0$, where $\vy_0$ serves as the prompt for the tracking model $f$, at the last timestamp, we treat $\vx_0^\dagger$ as an unlabeled frame without prompts (\ie, $\vy_0^\dagger=\varnothing$).
This design allows us to use $\vy_0$, the ground-truth label of $\vx_{0}^\dagger$, to supervise the fine-tuning of $f$---by minimizing the discrepancy between the predicted $\hat{\vy}_{0}^\dagger$ and ${\vy}_0$. The rationale is that if $f$ fails to track traits in the intermediate frames (\ie, $\vx_1$ and $\vx_1^\dagger$), it will not carry useful information for correctly segmenting  $\vx_{0}^\dagger$ (see~\cref{fig-6: CCL}).

We use a combination of binary cross entropy (BCE) loss and Dice loss for fine-tuning, both of which are commonly used in training segmentation models. 

\textbf{Implementation details.} 
We assume access to one labeled image $(\vx_0, \vy_0)$ and a set of unlabeled training images disjoint from the test images.
At each fine-tuning step, we sample $\vx_1$ from the unlabeled set and create a short palindrome-style cycle. We apply LoRA~\cite{hulora} to fine-tune the decoder and memory encoder of video segmentation models.

Since short palindrome-style cycles $\{\vx_0, \vx_1, \vx_1^\dagger, \vx_0^\dagger\}$ are used, the memory bank can retain the prompt $\vy_0$ until the closing frame, potentially making fine-tuning ineffective. To address this, we apply the following strategy.    
\begin{enumerate}
[nosep,topsep=4pt,parsep=0pt,partopsep=4pt, leftmargin=*]
\item During the forward pass, we reset the memory bank after processing  $\vx_1$, preventing it from carrying $\vy_0$ to $\vx_0^\dagger$.
\item To propagate tracking information from $\vx_1$ to $\vx_1^\dagger$, we use the predicted mask $\hat{\vy}_1$ from the former as the prompt for the latter. Notably, this prompt remains differentiable. 
\end{enumerate}
With this strategy, minimizing \OursL encourages $f$ to predict an accurate $\hat{\vy}_1$, ensuring that it propagates useful information for correctly segmenting ${\vx_0^\dagger}$ at the final timestamp to match the ground-truth $\vy_0$.

\subsection{Extension to Trait-Based Retrieval}
\label{ss:approach_retireval}

Beyond trait segmentation within the same species, \OursA can also be used to retrieve specimens exhibiting similar traits (\eg, the white band on the forewing or the orange tiger tails on the hindwing) across different species. Given a query image $\vx_0$ and a {target} trait $\vy_0\star$---a single channel in the original $\vy_0\in\{0, 1\}^{W\times H \times C}$---\OursA scores each image $\vx_i$ in the retrieval pool by
\begin{enumerate}
[nosep,topsep=4pt,parsep=0pt,partopsep=4pt, leftmargin=*]
\item creating a palindrome-style cycle $\{\vx_0, \vx_i, \vx_i^\dagger, \vx_0^\dagger\}$; 
\item using $\vy_0\star$ as the prompt and taking the forward pass introduced in~\cref{3.2. OC-CCl} to predict $\hat{\vy}_0^\dagger\star$; 
\item calculating the IoU between $\vy_0\star$ and $\hat{\vy}_0^\dagger\star$.
\end{enumerate}
Namely, if $\vx_i$ has the target trait, the trait mask $\vy_0\star$ should accurately propagate to $\vx_i$ and then propagate back to $\vx_0$.

\section{Experiment}
\label{s:exp}

\begin{table*}[t]
\begin{minipage}{0.6\textwidth}
    \centering
    \caption{\small \textbf{Specimen segmentation results (mIoU) and computational cost.} \OursA outperforms recent FSS methods as well as standard many-shot segmentation models trained on full data, while with much less inference computational cost.}
    \vskip-5pt
    \setlength{\tabcolsep}{3pt}
    \resizebox{\columnwidth}{!}{
    \begin{tabular}{@{}ccccccc@{}}
    \toprule
    \textbf{\# of Data} & \textbf{Model} & \textbf{Time (s)} & \textbf{Major~\cite{lawrence2024heliconius}} & \textbf{Minor~\cite{lawrence2024heliconius}} & \textbf{Fish~\cite{fishvistaData}} & \textbf{Beetle~\cite{Fluck2018_NEON_Beetle}} \\ 
    \toprule
    \multirow{4}{*}{One-Shot} & HDMNet~\cite{xue2024hdmnet} & 0.53$_{\pm 0.00}$
    & 4.2$_{\pm 0.8}$ & 4.0 & 2.4 & 6.1$_{\pm 2.0}$ \\
    & PFENet~\cite{tian2020pfenet} & 0.43$_{\pm 0.05}$ & 8.0$_{\pm 3.4}$ & 4.2 & 3.1 & 19.1$_{\pm 3.4}$ \\
    & VAT~\cite{hong2021VAT} & 0.39$_{\pm 0.03}$
    & 13.5$_{\pm 3.7}$ & 15.1 & 24.6 & 26.0$_{\pm 5.7}$ \\
    & SegGPT~\cite{wang2023seggpt} & 0.27$_{\pm 0.04}$ & 35.2$_{\pm 2.9}$ & 41.9 & 54.9 & 45.2$_{\pm 4.3}$ \\
    \midrule
    \multirow{3}{*}{One-Shot} & DEVA~\cite{cheng2023tracking} + \OursA & 0.13$_{\pm 0.04}$ & 73.1$_{\pm 4.0}$ & 68.6 & 50.8 & 39.0$_{\pm 6.9}$ \\
    & Cutie~\cite{cheng2024putting} + \OursA & 0.11$_{\pm 0.04}$ & 67.4$_{\pm 5.0}$ & 69.7 & 51.9 & 45.8$_{\pm 4.4}$ \\
    & SAM 2~\cite{ravi2024sam} + \OursA & 0.30$_{\pm 0.06}$ & 81.0$_{\pm 1.0}$ & 70.6 & 70.4 & 61.9$_{\pm 3.7}$ \\
    \midrule
    \multirow{2}{*}{\textcolor{gray}{Full}} & \textcolor{gray}{YOLOv8~\cite{Jocher_Jing_Chaurasia}} & 0.51$_{\pm 0.12}$ & \textcolor{gray}{71.1} & \textcolor{gray}{-} & \textcolor{gray}{-} & \textcolor{gray}{75.1} \\
    & \textcolor{gray}{Mask2Former~\cite{cheng2022masked}} & 0.56$_{\pm 0.07}$ & \textcolor{gray}{79.3} & \textcolor{gray}{-} & \textcolor{gray}{-} & \textcolor{gray}{83.2} \\
    \bottomrule
    \end{tabular}
    }
    \label{tab-1: main mIoU}
\end{minipage}
\hfill
\raisebox{0.6em}{
\begin{minipage}{0.38\linewidth}
\centering
\includegraphics[width=0.85\linewidth]{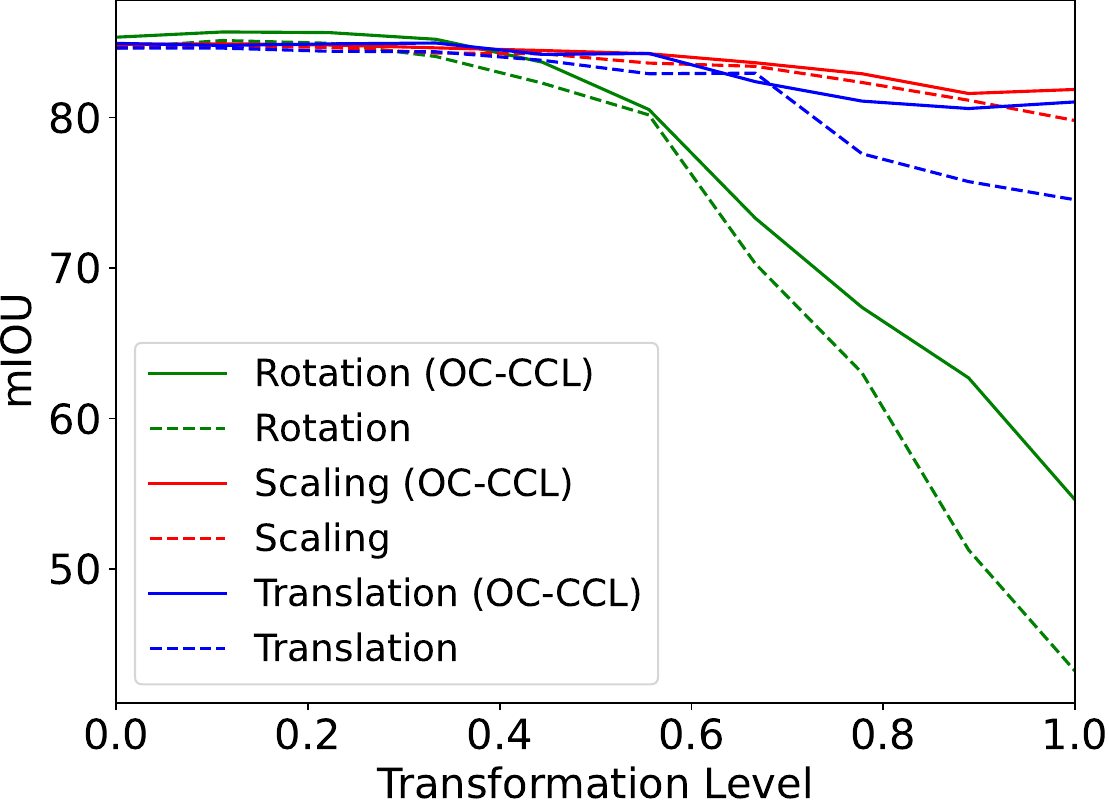}
\vskip-6pt
\captionof{figure}{\small \textbf{Fine-tuning results.} OC-CCL fine-tuning improves \OursA's robustness to transformation variations.}
\vskip-20pt
\label{fig-8: Breaking Factors}
\end{minipage}
}
\vskip -5pt
\end{table*}

\subsection{Experimental Setup} 

\textbf{Data.} We evaluate \OursA on three specimen data sources.
\begin{itemize}[nosep,topsep=4pt,parsep=0pt,partopsep=4pt, leftmargin=*]
\item \textbf{Butterfly:} We use the Cambridge Heliconius Collection \cite{lawrence2024heliconius}\footnote{Sources:~\cite{gabriela_montejo_kovacevich_2020_4289223,patricio_a_salazar_2020_4288311,montejo_kovacevich_2019_2677821,jiggins_2019_2682458,montejo_kovacevich_2019_2682669,montejo_kovacevich_2019_2684906,montejo_kovacevich_2019_2686762,montejo_kovacevich_2019_2702457,montejo_kovacevich_2019_2707828,montejo_kovacevich_2019_2714333,montejo_kovacevich_2019_2813153,montejo_kovacevich_2019_3082688,montejo_kovacevich_2021_5526257,warren_2019_2552371,warren_2019_2553977,gabriela_montejo_kovacevich_2019_3569598,gabriela_montejo_kovacevich_2020_4287444,gabriela_montejo_kovacevich_2020_4288250,gabriela_montejo_kovacevich_2020_4291095,jiggins_2019_2549524,jiggins_2019_2550097,warren_2019_2553501,salazar_2018_1748277,salazar_2019_2548678,salazar_2019_2735056,mattila_2019_2554218,mattila_2019_2555086,pinheiro_de_castro_2022_5561246,joana_i_meier_2020_4153502}. }, annotated in consultation with biologists and the field guide \cite{Heliconius}. 
Due to its long-tailed distribution (see \cref{fig-2: specimen}), we divide the dataset into Major and Minor parts. 
The Major part comprises the five largest subspecies, with $100$ specimens per subspecies for testing and the remaining $2,831$ samples for training on a subspecies basis. The Minor part consists of 146 subspecies with $2\sim 3$ hand-labeled samples for each of them. The training set cannot be constructed for the Minor part due to insufficient samples per subspecies.

\item \textbf{Fish}: We use the Fish-Vista dataset~\cite{fishvistaData}, containing specimens from over $1,900$ species.\footnote{This dataset is comprised of specimen images from various collections:~\cite{inhs, jfbm, ummz, umadison, fmnh, osum, fishair, Morphbank, glin, IDigBio, mabee2018phenoscape, edmunds2015phenoscape, mabee2012500}.} A subset ($1,573$ images across $474$ species) was labeled with $9$ expert-selected body parts; the labels are consistent across species. Similar to Butterfly Minor, there is no training set for Fish. 
\item \textbf{Beetle}: We use the individual image subset of the 2018 NEON-beetles dataset \cite{Fluck2018_NEON_Beetle}. We hand-labeled $180$ specimen images ($120$/$60$ for training/testing) with $5$ body parts. 
The data is challenging due to up to $90$-degree of body rotations, missing, and overlapped parts. 
\end{itemize}

We note that our task imposes challenges from its fine-grained nature and the vast diversity of species.
As such, standard few-shot datasets may not be ideal for development. 
Additional dataset details are provided in the Appendix.

\begin{table*}
    \centering
    \caption{\small \textbf{Opening-Closing Cycle-Consistent Loss and multi-shot results.} Applying \OursL and Multi-Shot inference on top of \OursA contributes to further performance improvement, surpassing SegGPT by a large margin.}
    \vskip -4pt
    \resizebox{\textwidth}{!}{
    \begin{tabular}{@{}cc|cc|cc|cc|cc@{}}
    \toprule
    \multicolumn{2}{c|}{\textbf{Modules}} & \multicolumn{2}{c|}{\textbf{SegGPT}~\cite{wang2023seggpt}} & \multicolumn{2}{c|}{\textbf{DEVA}~\cite{cheng2023tracking} + \textbf{\OursA}} & \multicolumn{2}{c|}{\textbf{Cutie}~\cite{cheng2024putting} + \textbf{\OursA}} & \multicolumn{2}{c}{\textbf{SAM 2}~\cite{ravi2024sam} + \textbf{\OursA}} \\
    \textbf{\OursL} & \textbf{Five-Shot} & \textbf{Major}~\cite{lawrence2024heliconius} &\textbf{Beetle}~\cite{Fluck2018_NEON_Beetle} & \textbf{Major}~\cite{lawrence2024heliconius} &\textbf{Beetle}~\cite{Fluck2018_NEON_Beetle} & \textbf{Major}~\cite{lawrence2024heliconius} &\textbf{Beetle}~\cite{Fluck2018_NEON_Beetle} & \textbf{Major}~\cite{lawrence2024heliconius} &\textbf{Beetle}~\cite{Fluck2018_NEON_Beetle} \\
    \midrule
    & & 35.2$_{\pm 2.9}$ & 45.2$_{\pm 4.3}$ & 73.1$_{\pm 4.0}$ & 39.0$_{\pm 6.9}$ & 67.4$_{\pm 5.0}$ & 45.8$_{\pm 4.4}$ & 81.0$_{\pm 1.0}$ & 61.9$_{\pm 3.7}$\\
    \checkmark & & - & - & 74.2$_{\pm 2.5}$ & 48.4$_{\pm 6.7}$ & 77.7$_{\pm 2.8}$ & 48.7$_{\pm 2.2}$ & 81.2$_{\pm 1.0}$ & 65.2$_{\pm 3.3}$ \\
    & \checkmark & 42.3$_{\pm 0.3}$ & 70.3$_{\pm 2.7}$ & 83.2$_{\pm 1.9}$ & 65.5$_{\pm 2.4}$ & 83.4$_{\pm 1.7}$ & 67.2$_{\pm 2.0}$ & 83.4$_{\pm 0.4}$ & 74.2$_{\pm 2.9}$\\
    \checkmark & \checkmark & - & - & 83.4$_{\pm 0.7}$ & 66.9$_{\pm 1.3}$ & 84.7$_{\pm 1.3}$ & 67.6$_{\pm 1.1}$& 83.9$_{\pm 0.2}$ &75.6$_{\pm 1.5}$\\
    \bottomrule
    \end{tabular}
    }
\label{tab-2: occcl}
\vskip -13pt
\end{table*}

\textbf{Evaluation metric.}
We use mean IoU (mIoU) as the main metric, averaged over traits in images. For FSS methods (including \OursA) in a one-shot setting, one labeled image (with canonical shapes and visually clear traits) is sampled as the reference for the test set. We report averaged mIoU over $20$ runs with standard deviation, unless stated otherwise.

\textbf{Implementation details of \OursA.} 
We apply \OursA to three video segmentation models, DEVA~\cite{cheng2023tracking}, Cutie~\cite{cheng2024putting}, and SAM 2~\cite{ravi2024sam}, using the official pre-trained models. 

\textbf{Baselines.}
We consider four representative few-shot segmentation (FSS) methods, PFENet \cite{tian2020pfenet}, VAT \cite{hong2021VAT}, HDMNet \cite{xue2024hdmnet}, and SegGPT \cite{wang2023seggpt}. 
We also apply two representative many-shot instance segmentation algorithms, YOLOv8 \cite{Jocher_Jing_Chaurasia} and Mask2Former \cite{cheng2022masked}, whenever we have sufficient training samples.
The Appendix presents more baseline settings and comparisons.

\begin{table*}[!t]
\begin{minipage}{0.39\linewidth}
    \vskip5pt
    \centering
    \small
    \captionof{table}{\small\textbf{Instance segmentation results on CUB.} Our method \OursA achieves similar results as SOTA FSS methods on object instance segmentation.}
    \resizebox{\columnwidth}{!}{
    \begin{tabular}{@{}lcc@{}}
        \toprule
        \textbf{Model} & \textbf{One-Shot} & \textbf{Five-Shot} \\ \toprule
        HDMNet~\cite{xue2024hdmnet}    & 65.8$_{\pm 1.4}$ & 66.3$_{\pm 1.4}$ \\
        PFENet~\cite{tian2020pfenet}   & 72.2$_{\pm 0.6}$ & 73.1$_{\pm 0.4}$\\
        VAT~\cite{hong2021VAT}         & 83.4$_{\pm 1.0}$ & 85.3$_{\pm 0.7}$\\
        SegGPT~\cite{wang2023seggpt}   & 51.3$_{\pm 2.9}$ & 78.8$_{\pm 1.4}$\\
        SAM 2 + \OursA~(\textbf{ours}) & 71.1$_{\pm 1.3}$ & 77.9$_{\pm 0.8}$\\
        SAM 2 + \OursA+ OC-CCL~(\textbf{ours}) & 77.8$_{\pm 1.1}$ & 79.6$_{\pm 0.4}$\\ \bottomrule
    \end{tabular} 
    }
    \label{tab-4: Instance Result}
\end{minipage}
\hfill
\begin{minipage}{0.59\linewidth}
\includegraphics[width=1.0\linewidth]{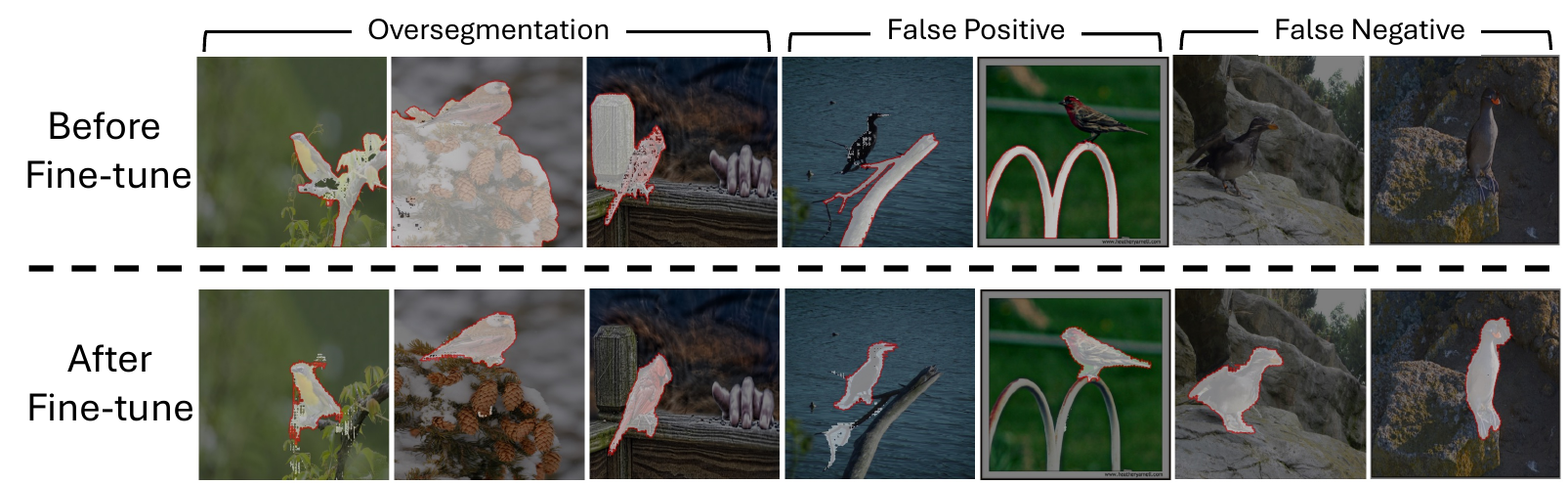}
\vskip-5pt
\captionof{figure}{\small \textbf{Before vs. after OC-CCL fine-tuning.} 
Fine-tuning with OC-CCL notably improves \OursA with merely one labeled training example.}
\label{fig-9: birds fine-tune}
\end{minipage}
\vskip -8pt
\end{table*}

\subsection{Main Result}

\begin{figure}[t]
  \centering
\begin{subfigure}[c]{0.6\textwidth}
  \includegraphics[width=\linewidth]{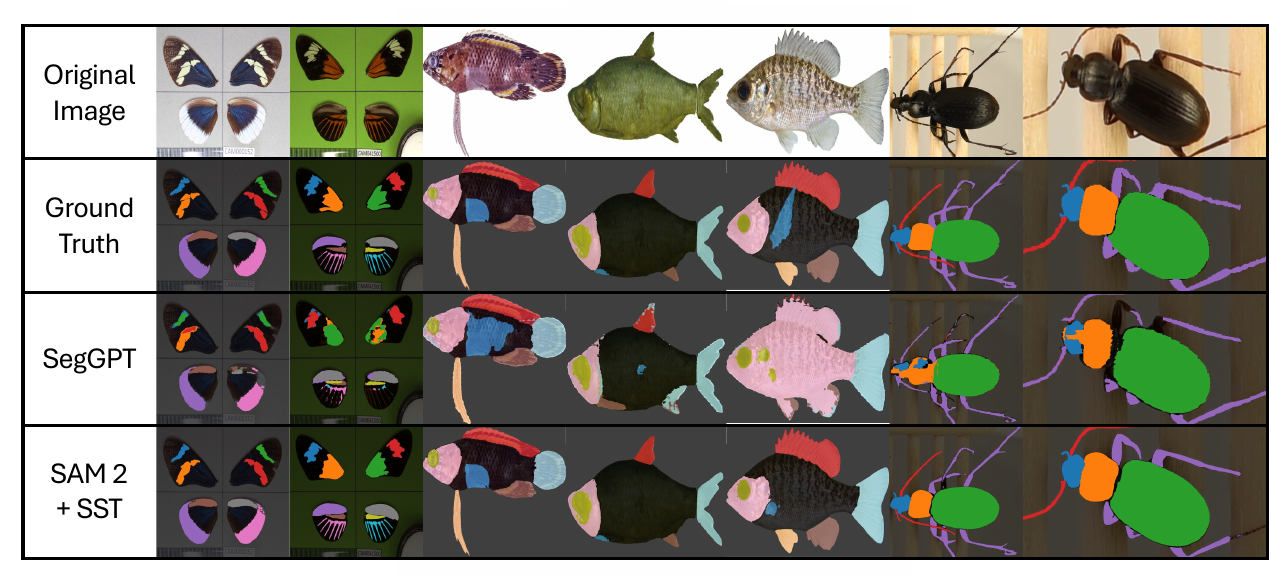}
  \caption{}
  \label{fig-7: qualitative}
  \vspace{-10pt}
\end{subfigure}
\hfill
\begin{subfigure}[c]{0.38\textwidth}
    \centering
    \includegraphics[width=\linewidth]{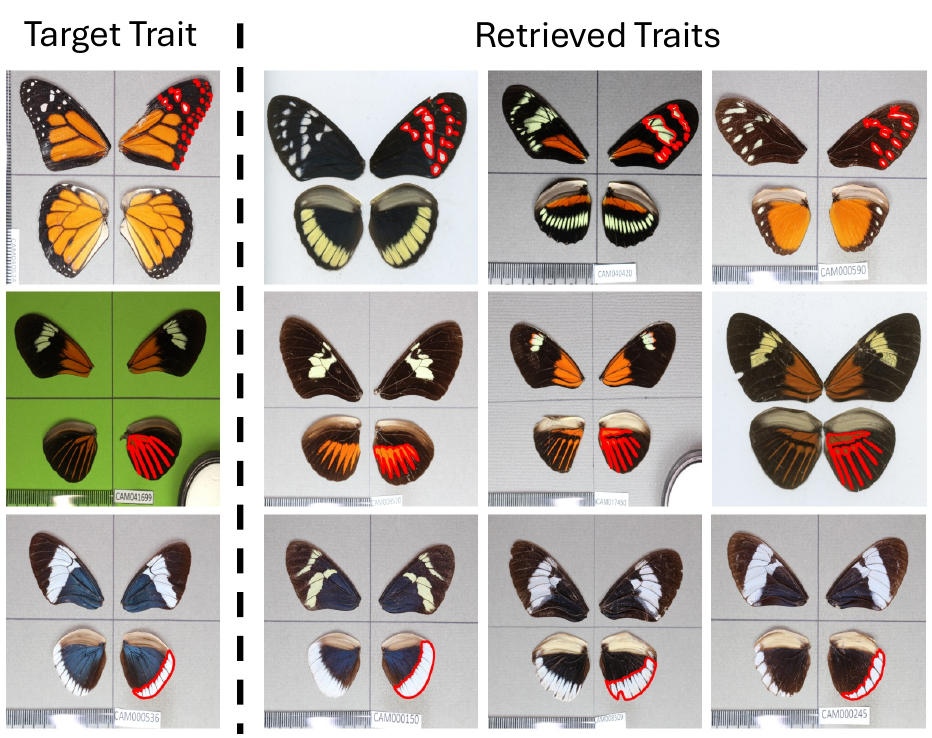}
    \caption{}
    \label{fig-10: trait retrieval}
\end{subfigure}
\vspace{-5pt}
\caption{\small \textbf{(a) Qualitative results:} Trait segmentation of \OursA (with SAM 2) vs.\ SegGPT on butterfly, fish, and beetle data. \textbf{(b) Trait-based retrieval.} We can retrieve different butterfly subspecies with visually consistent traits to the specified target (marked in red).}
\vspace{-10pt}
\end{figure}

\textbf{Specimen segmentation results.}
We evaluate \OursA on Butterfly~\cite{lawrence2024heliconius} (Major, Minor), Fish~\cite{fishvistaData}, and Beetle~\cite{Fluck2018_NEON_Beetle} datasets. 
On Butterfly Major and Beetle with sufficient training data, we train Mask2Former~\cite{cheng2022masked} and YOLOv8 \cite{Jocher_Jing_Chaurasia} models using all the training samples per (sub)species as many-shot baselines. For FSS algorithms and \OursA, we consider a one-shot setting.
On the remaining datasets (Butterfly Minor, Fish), where samples per (sub)species are limited, we evaluate only in the one-shot setting in a leave-one-test-sample-out manner; standard deviation is not reported.

As shown in~\cref{tab-1: main mIoU}, \OursA outperforms existing FSS algorithms across most datasets using various video segmentation models. Especially on Butterfly Minor, \OursA achieves a margin of at least $26.7$ mIoU over SegGPT~\cite{wang2023seggpt}. Remarkably, \OursA based on SAM 2~\cite{ravi2024sam} even surpasses many-shot algorithms trained with at least $150$ samples per subspecies on Butterfly Major. 
As evidenced in~\cref{fig-7: qualitative}, \OursA offers more accurate segmentation results than SegGPT~\cite{wang2023seggpt}. 

The superior performance of \OursA can be primarily attributed to its alignment with the task. Fine-grained specimen segmentation is inherently more challenging than the tasks for which existing FSS models were designed. 
However, specimen images exhibit strong interdependencies, even when captured non-sequentially, making video segmentation particularly well-suited for this problem.

\textbf{Opening-Closing Cycle-Consistent Loss.}
Given sufficient data from Butterfly Major~\cite{lawrence2024heliconius} and Beetle~\cite{Fluck2018_NEON_Beetle}, we further fine-tune the pre-trained models with \OursL.
For each species, one labeled image and the other unlabeled images from the training set are used to construct the palindrome cycles. 
We evaluate the effectiveness of \OursL in~\cref{tab-2: occcl}, where it is applied to all three tracking models (the second row), with consistent improvement across all tasks. 

\textbf{Multi-shot inference.}
We have investigated the one-shot case where only one labeled sample is leveraged in~\cref{tab-1: main mIoU}. Given more labeled samples, \OursA also supports multi-shot inference with more comprehensive information.
As shown in~\cref{tab-2: occcl}, using multiple reference frames enables \OursA to outperform YOLOv8~\cite{Jocher_Jing_Chaurasia} and Mask2Former~\cite{cheng2022masked}, while boosting Cutie~\cite{cheng2024putting} to performance levels comparable with SAM 2~\cite{ravi2024sam} on Butterfly Major. 
Moreover, combining multi-shot inference with \OursL yields considerable performance improvement across all metrics. 
Please refer to the Appendix for more details of the multi-shot setting. 

\subsection{Analysis}

\textbf{Out-of-distribution (OOD) robustness.}
\label{ss:ood-robustness}
The actual application of \OursA on specimen images might encounter OOD cases, where the specimens are not captured in standard views. 
That is, the images might be subjected to rotation, translation, or scaling. 
Accordingly, we manually apply these transformations to the Butterfly test set to create OOD cases. 
We define transformation levels from $0.0$ to $1.0$, corresponding to no transformation and the largest degree of transformation we apply, respectively.
At level $1.0$, we randomly rotate the image between $-90$° and $90$°, translate it up to $60\%$ of its height or width, and scale it down to $50\%$ of its original size.
We found that \OursA tends to lose track of the fine-grained details after a certain level of transformations, likely due to the absence of such huge variations (between consecutive frames) in the pre-training data. 

To address this, we use \OursL to fine-tune the model in a one-shot setting for each test image. \OursL consistently improves \OursA's robustness as seen in \cref{fig-8: Breaking Factors}. In the extreme rotation cases (the right end of the figure), we boost the mIoU from $40\%$ to over $50\%$, a more than $10\%$ gain.

\subsection{Extension and Further Exploration}
\label{ss:further_explore}

\begin{wrapfigure}{r}{0.6\textwidth}
  \vspace{-10pt}
  \begin{minipage}{\linewidth}
    \centering
    \small
    \captionof{table}{\textbf{Segmentation results.} Model performance comparison on CelebAMask-HQ and MRBrainS.}
    \label{tab:facial-mri-performances}
    \resizebox{\linewidth}{!}{%
      \begin{tabular}{@{}lcccc@{}}
        \toprule
        \textbf{Dataset} & \textbf{SegGPT} & \textbf{DEVA+\OursA} & \textbf{Cutie+\OursA} & \textbf{SAM~2+\OursA} \\
        \midrule
        CelebAMask-HQ & 58.9$_{\pm1.7}$ & 62.3$_{\pm4.3}$ & 52.6$_{\pm5.6}$ & 73.2$_{\pm2.2}$ \\
        MRBrainS      & 42.7$_{\pm2.7}$ & 46.6$_{\pm4.3}$ & 51.4$_{\pm2.9}$ & 52.6$_{\pm4.1}$ \\
        \bottomrule
      \end{tabular}%
    }
  \end{minipage}
  \vspace{-5pt}
\end{wrapfigure}

\textbf{Instance segmentation.}
Besides fine-grained specimen segmentation, \OursA can also be applied to standard object instance segmentation on images taken in the wild. We use the CUB-200-2011 dataset \cite{wah2011caltech} to demonstrate such a capability. Given one/five random bird images and their segmentation masks, we examine if FSS algorithms can segment all 200 bird species from the remaining images. 
The results are shown in~\cref{tab-4: Instance Result}.
As whole object instance segmentation is the original problem domain for most of the compared methods, they show much better results than~\cref{tab-1: main mIoU} and~\cref{tab-2: occcl}. Even with large variations from image to image, \OursA still achieves a competitive segmentation performance across bird images. Furthermore, fine-tuning \OursA with training images using \OursL again shows significant improvement in segmentation quality. We closely analyze the object instances that originally fail to be correctly segmented by \OursA and categorize them into 3 failure cases: \textit{Oversegmentation}, where \OursA correctly segments out the object along with some extra neighboring backgrounds; \textit{False Positive}, where \OursA falsely segments out an irrelevant object; and \textit{False Negative}, where \OursA completely fails to segment anything from the picture. As shown in the bottom row of \cref{fig-9: birds fine-tune}, without using any ground truth masks for most training images, fine-tuning with \OursL helps substantially mitigate these issues.

\textbf{Trait-based retrieval.}
As mentioned in \cref{ss:approach_retireval}, given a target trait, our method can use the reconstruction IoU to find images with similar traits. \cref{fig-10: trait retrieval} shows that \OursA+ \OursL faithfully retrieves images with similar corresponding traits, which can be useful for studying similar subspecies.
For more experimental results and discussions, please see the Appendix.

\textbf{Further exploration.}
\OursA leverages the inherent structural dependency between static specimen images to enable tracking-based segmentation. Similar dependencies exist in other domains, such as facial and medical images. 
\cref{tab:facial-mri-performances} suggests superior segmentation performance of \OursA over the other FSS methods, indicating that \OursA successfully captures the underlying relationships that previous FSS methods fail to exploit. 
Please refer to the Appendix for more detailed results.

\section{Conclusion and Discussion}
\label{sec: conclusion}

We introduce Static Segmentation by Tracking (\OursA), a label-efficient approach for fine-grained specimen image segmentation. By applying a tracking algorithm like SAM~2 to non-sequential specimen images, \OursA achieves remarkable trait segmentation using only a single labeled image. Further analysis reveals that \OursA extends beyond specimen images, successfully segmenting animal instances in the wild. Additionally, it enables trait-level retrieval, identifying species with similar traits and patterns.

While our main use case is specimen images, this does not imply that our scope and applicability are ``limited.''
First, specimens are a major resource for biologists to understand organisms, and a vast amount of specimens have yet to be digitized and analyzed. 
Machine learning techniques are enabling efficient data processing, saving excessive manual efforts required for dealing with specimen images. 
Second, in addition to specimens, object-centric images with canonical poses are a common image source in various scientific fields such as MRI and CT scans. The method can be further extended for broader application. 
Third, while specimen images may seem easier to handle at first glance due to their object-centric nature and plain backgrounds (compared to natural images like those in MS-COCO~\cite{lin2014microsoft}), our experiments demonstrate that segmenting fine-grained traits from them is non-trivial, particularly in a few-shot setting.
In summary, our paper contributes not only to the computer vision community (\eg, by promoting a rarely studied but challenging task, providing data for benchmarking, and offering a novel approach) but also to other scientific communities (\eg, by facilitating the measurement of traits).

\clearpage
{
    \small
    \bibliographystyle{ieeenat_fullname}
    \bibliography{main}
}
\clearpage
\setcounter{page}{1}

{
\centering
\Large
  \textbf{\emph{Static Segmentation by Tracking}:\\
  A Label-Efficient Approach for Fine-Grained Specimen Image Segmentation}
  \\[0.5em]
Appendix\\[1em]
}
\appendix

\renewcommand{\thesection}{A\arabic{section}}  
\renewcommand{\thetable}{A\arabic{table}}  
\renewcommand{\thefigure}{A\arabic{figure}}
\setcounter{table}{0}
\setcounter{figure}{0}
\setcounter{equation}{0}
\setcounter{section}{0}

\begin{table}[h]
\centering
\centering
\begin{minipage}{\columnwidth}
\centering
\caption{\textbf{Dataset statistics} for Butterfly~\cite{lawrence2024heliconius}, Fish~\cite{fishvistaData}, and Beetle~\cite{Fluck2018_NEON_Beetle} dataset.}
\small
\begin{tabular}{@{}lcccc@{}}
\toprule
                   & \multicolumn{2}{c}{\textbf{Butterfly}} & \multirow{2}{*}{\textbf{Fish}} & \multirow{2}{*}{\textbf{Beetle}} \\ \cmidrule(lr){2-3}
                   & \textbf{Major} & \textbf{Minor} &  &  \\ \midrule
\textbf{\# of Classes} & 5                   & 146                    & 474                           & 12                               \\
\textbf{Total Train}   & 2,831                & -                    & -                              & 120                              \\
\textbf{Total Test}    & 500                 & 313                    & 1,573                               & 60                               \\ \bottomrule
\end{tabular}
\label{tab:grouped-species-stats}
\end{minipage}

\end{table}

\section{Dataset Statistics and Evaluation Details}

We put the general statistics for Butterfly~\cite{lawrence2024heliconius}, Fish~\cite{fishvistaData}, and Beetle~\cite{Fluck2018_NEON_Beetle} datasets in \cref{tab:grouped-species-stats}.

\subsection{Butterfly.} The Cambridge Butterfly~\cite{lawrence2024heliconius} dataset includes $151$ butterfly subspecies of the Heliconius genus; each has $4\sim 14$ distinctive traits to tell itself apart from others. Example traits include the tiger tails on the hindwings and white bands on the forewings; some have quite complex, disconnected shapes. 
Across specimens of the same subspecies, the mask IDs are consistent. An algorithm needs to segment them and also label each with an ID.

We split the dataset into two parts, Major and Minor, based on the available data sample for each subspecies. 
The Major part has 5 different subspecies, with 2,831 semi-automatically labeled training samples and 500 hand-labeled test samples in total. 
The semi-automatic approach: we used \OursA to propagate masks, followed by human inspection. \emph{Samples with unsatisfactory masks were then hand-labeled.}
We take all $2,831$ training data samples to train standard segmentation models, Mask2Former~\cite{cheng2022masked} and YOLOv8~\cite{Jocher_Jing_Chaurasia}, for each subspecies. 
To evaluate few-shot segmentation models, we sample a specified number of random specimens from the train set for each subspecies, and evaluate the performance on the corresponding test data. 

The Minor part has $146$ subspecies with $313$ hand-labeled test images in total.
As there is insufficient data for each subspecies to construct a training set, Minor is intended for the one-shot segmentation task. For this part, we only test on few-shot segmentation models in the same fashion as we evaluate the Major subspecies.

\subsection{Fish.} The Fish-Vista~\cite{fishvistaData} dataset has $474$ different species of fish, containing $1,573$ samples in total. All species share a common set of 9 segmentation classes (\eg, head, eye, tail, adipose fin, caudal fin, etc.). As there are a limited number of samples per species, we do not run many-shot model training or \OursL fine-tuning on Fish. For few-shot segmentation, we use $1$ sample from each species as reference, and predict the segmentation masks of the other samples.

\subsection{Beetle.} The Beetle~\cite{Fluck2018_NEON_Beetle} dataset consists of beetles of 12 different species.
Each beetle species shares 5 common segmentation classes: head, pronotum, elytra, antenna, and legs. The antennae and leg parts of beetles are quite challenging, with complex, sharp, and thin shapes. 
As the beetle species share similar visual traits, we always apply a universal model across all the species. 
For each species, we hand-label 15 images in total, taking 10 as the training set and 5 as the test set. 
For the many-shot instance segmentation models, we train one single model on all training data across species.
For few-shot segmentation methods and \OursA, we sample one example from the 120 training samples and test the segmentation quality on all 60 test samples across species.

\subsection{Remark.}
We emphasize that while we used SAM~2 to assist in data annotation, \emph{humans} inspected the results and re-labeled them when necessary. Additionally, the test set was entirely labeled by \emph{humans}. Consequently, a well-trained model, such as Mask2Former \cite{cheng2022masked}, given sufficient labeled data, could surpass
\OursA in performance.

\section{Implementation Details}

\subsection{Many-Shot Methods}
Existing many-shot methods usually rely on sufficient training data to obtain ideal segmentation performance. As shown in~\cref{fig-1: performance comparison}, YOLOv8~\cite{Jocher_Jing_Chaurasia} and Mask2Former~\cite{cheng2022masked} require more than $1,000$ training samples to achieve comparable performance with \OursA. That means the many-shot methods cannot work as expected for Butterfly Minor~\cite{lawrence2024heliconius} and Fish~\cite{fishvistaData}, where insufficient data is provided within each (sub)species. Therefore, we only train many-shot models for Butterfly Major~\cite{lawrence2024heliconius} and Beetle~\cite{Fluck2018_NEON_Beetle} datasets. 
More specifically, for Butterfly Major~\cite{lawrence2024heliconius}, we train independent segmentation models for each subspecies, while for Beetle~\cite{Fluck2018_NEON_Beetle}, only one model is trained for the whole dataset. 
The training is conducted following standard hyper-parameter settings of the adopted methods. 

\subsection{\OursL Fine-tuning}
Similar to many-shot learning methods, \OursL fine-tuning cannot be applied to Butterfly Minor~\cite{lawrence2024heliconius} or Fish~\cite{fishvistaData} due to insufficient samples with each (sub)species. 
For Beetle~\cite{Fluck2018_NEON_Beetle} and Butterfly Major~\cite{lawrence2024heliconius}, we follow the default LoRA initialization and learning rate scheduling for LoRA~\cite{hulora}, and train it for one epoch. 
More specifically, we replace all linear projection layers and MLP layers in the memory encoder and mask decoder of the video segmentation models with LoRA linear layers, using $r=32$ and $\alpha=64$. We take one fixed labeled image as $\vx_0$, and iteratively sample $\vx_1$ from the unlabeled training set to create multiple short palindrome cycles, as introduced in \cref{3.2. OC-CCl}. For Beetle~\cite{Fluck2018_NEON_Beetle}, due to the inherently high variation in beetle orientations, we introduce a small degree of random 2D rotation augmentation on the labeled example during the fine-tuning stage to enhance the robustness of the model towards rotation OOD scenarios.

\begin{table}[h]
    \centering
    \small
    \caption{\textbf{OC-CCL fine-tuning time} on different datasets for one epoch (minutes).}
    \begin{tabular}{lcc}
        \toprule
        \textbf{Model} & \textbf{Major}~\cite{lawrence2024heliconius} & \textbf{Beetle}~\cite{Fluck2018_NEON_Beetle} \\  
        \midrule
        DEVA~\cite{cheng2023tracking} & 10 & 5 \\  
        Cutie~\cite{cheng2024putting} & 10 & 4 \\  
        SAM 2~\cite{ravi2024sam} & 6  & 2 \\  
        \bottomrule
    \end{tabular}
    \label{tab:training_time}
\end{table}

As illustrated in~\cref{tab:training_time}, the fine-tuning does not require excessive time for the adopted datasets. It enhances the application value for \OursA on real-world scenarios. 

\subsection{One-Shot Segmentation Setting}
For the Butterfly Major~\cite{lawrence2024heliconius} and Beetle~\cite{Fluck2018_NEON_Beetle} datasets, we perform 20 runs of one-shot segmentation experiments to ensure a fair and stable comparison. In each run, we sample a different labeled image from the training set as the reference frame and evaluate across all testing set examples. The average mIoU is then calculated across all runs with the standard deviation reported.

For Fish~\cite{fishvistaData} and Butterfly Minor~\cite{lawrence2024heliconius}, where there is insufficient data to sample 20 reference images, we conduct only one single run. Specifically, we use one labeled example as the reference and perform inference on the remaining data within the (sub)species.

\subsection{Multi-Shot Setting}
Similar to the other few-shot segmentation algorithms, \OursA can also benefit under the multi-shot setting. 
Given more labeled images, video segmentation models leverage them as multiple reference frames to better capture information for the subsequent prediction steps. 
For 5-shot evaluation, we adopt a naive design, where five labeled image mask pairs are put in the beginning of the video sequence, where a new sequence is created independently for each of the incoming images. In other words, given labeled reference pairs $(\vx_0,\vy_0)$, $(\vx_1,\vy_1)$, $(\vx_2,\vy_2)$, $(\vx_3,\vy_3)$, $(\vx_4,\vy_4)$ and unlabeled target images $\{\vx_5,\dots,\vx_N\}$, we create sequences $\{\vx_0,\dots,\vx_4,\vx_5\},\dots,\{\vx_0,\dots,\vx_4,\vx_N\}$ and apply \OursA to each of them. 
We also aim to design a better multi-shot mechanism for~\OursA as a future work.

\section{More Baselines}
In addition to FSS and many-shot instance segmentation models, we also adopt a naive ``copy and paste'' method and optical flow as supplementary baselines. 

\begin{table}[h]
    \centering
    \small
    \caption{\textbf{Segmentation results} across different datasets using copy baseline and optical flow.}
        \begin{tabular}{lccccc}
            \toprule
            & \textbf{Major}~\cite{lawrence2024heliconius} & \textbf{Minor~\cite{lawrence2024heliconius}} & \textbf{Fish~\cite{fishvistaData}} & \textbf{Beetle~\cite{Fluck2018_NEON_Beetle}} \\
            \midrule
            Copy & 20.9$_{\pm 10.9}$ & 36.1 & 10.6 & 13.2$_{\pm 6.7}$ \\
            Optical Flow & 24.9$_{\pm 11.9}$ & 36.2 & 5.7 & 16.5$_{\pm 4.8}$ \\
            \midrule
    DEVA~\cite{cheng2023tracking} + \OursA & 73.1$_{\pm 4.0}$ & 68.6 & 50.8 & 39.0$_{\pm 6.9}$ \\
    Cutie~\cite{cheng2024putting} + \OursA & 67.4$_{\pm 5.0}$ & 69.7 & 51.9 & 45.8$_{\pm 4.4}$ \\
    SAM 2~\cite{ravi2024sam} + \OursA & 81.0$_{\pm 1.0}$ & 70.6 & 70.4 & 61.9$_{\pm 3.7}$\\
            \bottomrule
        \end{tabular}
    \label{tab:supp-baselines}
\end{table}

\subsection{Copy Baseline.}
Given the highly aligned nature, an intuitive way to address the fine-grained specimen image segmentation problem is simply copying the segmentation masks from the reference image. 
We evaluate the copy baseline in~\cref{tab:supp-baselines}. While the naive solution achieves certain mIoU performance by chance, we demonstrate that on all three video segmentation models, \OursA yields much stronger results. 
It also indicates that although the problem seems straightforward, it takes non-trivial efforts to achieve practical results. 

\subsection{Optical Flow.}
Another intuitive solution to this problem is optical flow, as it's a fundamental way to calculate the pixel mapping from one image to the other. This technique is widely used in the medical image registration and video tracking field, which is similar to our task. To implement this, we first perform dense optical flow to calculate the per-pixel mapping from the labeled to unlabeled images. We then send the segmentation masks through the same mapping to ``map'' the masks to the corresponding region on the target image. As shown in \cref{tab:supp-baselines}, there is limited improvement using optical flow compared with the copy baseline. It further demonstrates that specimen trait segmentation cannot be easily solved through existing techniques.

\section{Discussion.}
\subsection{Fairness of Different Baselines.}
We note that different compared methods may vary in their parameters, pre-trained data, training and inference time, etc. Aligning them completely is challenging, especially since \OursA uses video models, which fundamentally differs from an image model in several aspects. 
That said, we use video models not for their training data, GPUs, or size, but for their video segmentation capability, which unexpectedly aligns with our problem.

\subsection{Images with Multiple Specimens.}
We assume that each image contains a single specimen instance. If an image contains multiple specimens, object detectors like Grounding DINO~\cite{liu2023grounding} can be applied beforehand to separate them.

\subsection{Effectiveness of Video Segmentation Models.}
We view this as an emergent property of models trained for video segmentation---they can track and segment taxonomically related species across non-sequential, independently captured photographs.

\section{Additional Analysis on \OursA}

\subsection{Computational Cost}
The inference stage of \OursA involves mask propagation across images, which might raise concerns about the computational cost. 
In~\cref{tab-1: main mIoU}, we report the time each method requires to process a single instance on one NVIDIA A100 GPU with 40 GB of VRAM.
\OursA demonstrates greater efficiency than most of the compared methods, confirming its practical applicability.

\subsection{Detailed Results on Face and MRI Images}

\begin{table}[h]
    \centering
    \small
    \caption{\textbf{Segmentation results} of different models across facial features on CelebAMask-HQ~\cite{CelebAMask-HQ}.}
        \begin{tabular}{@{}lcccc@{}}
            \toprule
            & \textbf{Overall} & \textbf{Eyes} & \textbf{Nose} & \textbf{Mouth} \\
            \midrule
            SegGPT~\cite{wang2023seggpt} & 58.9$_{\pm 1.7}$ & 46.7$_{\pm 2.6}$ & 68.7$_{\pm 3.3}$ & 61.3$_{\pm 5.7}$ \\
            DEVA~\cite{cheng2023tracking} + \OursA & 62.3$_{\pm 4.3}$ & 53.2$_{\pm 5.6}$ & 64.1$_{\pm 8.3}$ & 69.7$_{\pm 5.5}$ \\
            Cutie~\cite{cheng2024putting} + \OursA & 52.6$_{\pm 5.6}$ & 54.6$_{\pm 6.6}$ & 46.9$_{\pm 14.2}$ & 56.4$_{\pm 5.0}$ \\
            SAM 2~\cite{ravi2024sam} + \OursA & 73.2$_{\pm 2.2}$ & 70.3$_{\pm 1.4}$ & 64.2$_{\pm 5.3}$ & 81.8$_{\pm 2.2}$ \\
            \bottomrule
        \end{tabular}
    \label{tab:facial_segmentation}
\end{table}

\begin{table}[h]
    \centering
    \caption{\textbf{Segmentation results} of different models across brain tissue types on MRBrainS~\cite{mendrik2015mrbrains}.}
    \label{tab-6:Celeba-Mask}
    \setlength{\tabcolsep}{3pt}
        \begin{tabular}{@{}lcccc@{}}
        \toprule
        & \textbf{Overall} & \textbf{Cerebrospinal Fluid} & \textbf{Gray Matter} & \textbf{White Matter} \\
        \midrule
        SegGPT~\cite{wang2023seggpt} & 42.7$_{\pm 2.7}$ & 39.1$_{\pm 1.7}$ & 45.6$_{\pm 1.1}$ & 43.4$_{\pm 1.2}$ \\
        DEVA~\cite{cheng2023tracking} + \OursA & 46.6$_{\pm 4.3}$ & 42.0$_{\pm 3.7}$ & 52.3$_{\pm 2.1}$ & 45.6$_{\pm 2.0}$ \\
        Cutie~\cite{cheng2024putting} + \OursA & 51.4$_{\pm 2.9}$ & 49.8$_{\pm 1.9}$ & 55.5$_{\pm 2.5}$ & 48.8$_{\pm 2.4}$ \\
        SAM 2~\cite{ravi2024sam} + \OursA & 52.6$_{\pm 4.1}$ & 48.1$_{\pm 3.1}$ & 51.6$_{\pm 3.5}$ & 58.1$_{\pm 1.9}$ \\
        \bottomrule
    \end{tabular}
\end{table}

We report the detailed results on CelebAMask-HQ~\cite{CelebAMask-HQ} and MRBrainS~\cite{mendrik2015mrbrains} for different segmentation targets in~\cref{tab-6:Celeba-Mask} and~\cref{tab:facial_segmentation}, respectively. \OursA yields better performance than SegGPT~\cite{wang2023seggpt} on most of the tasks. 

\subsection{Analysis on Inference Variants}
\label{ordering_supp}
In the main paper, we evaluate our method using a single test image at a time to compare it against other few-shot segmentation models, ensuring a fair comparison. However, as mentioned in Section 3.3, it is also possible to concatenate all test images and process them together using \OursA. We observe that using random orders does not significantly degrade performance, though a more carefully designed algorithm could make \OursA more stable. 

\begin{figure}[t]
    \centering
    \vskip-10pt
    \includegraphics[width=0.5\columnwidth]{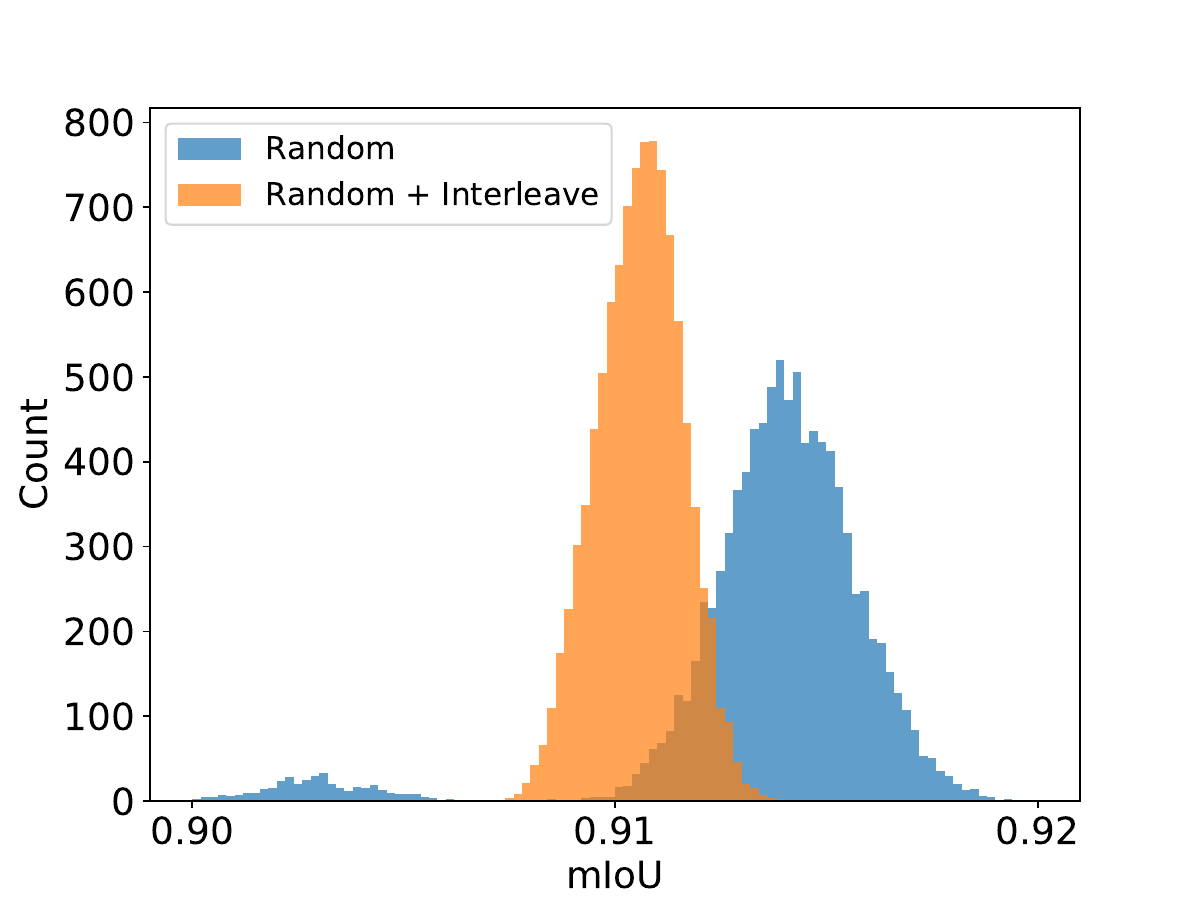}
    \vskip-10pt
    \caption{\small \textbf{\OursA performance on different inference variants.} Optimizing the frame ordering improves the stability when using long video sequences for inference. 
    }
    \label{fig-s1: sst ordering}
    \vskip -8pt
\end{figure}

To demonstrate this, we conduct a toy experiment on the lativitta subspecies from the Butterfly~\cite{lawrence2024heliconius} dataset. We first randomly sample $10$ butterflies from the test set and generate $10,000$ unique orderings for the $10$ images. We then put all ten images in a sequence based on each ordering and evaluate \OursA after propagating through the entire sequence. We average the mIoU performance across each ordering and plot it against a histogram as shown in the blue part of~\cref{fig-s1: sst ordering}. 
There appear to be two peaks in the distribution, but the overall influence is limited (mIoU from 90\% to 92\%).
We then experiment with a slightly improved ordering strategy by interleaving each test image with the reference image, so that the reference information can be retained even if some frames lose the track. As shown in the orange part of~\cref{fig-s1: sst ordering},
although random ordering only has a limited influence on the performance, interleaving the test images further reduces the standard deviation in mIoU. Thus, we conclude that finding the optimal sequence can indeed help with the stability of video inference, and we plan to explore the ordering design as a future work.

    \begin{figure}[t]
        \centering
        \includegraphics[width=0.8\columnwidth]{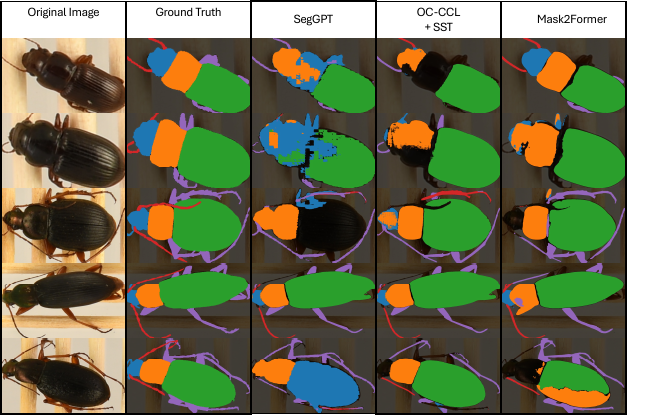}
        \caption{\small \textbf{Qualitative results on Beetle~\cite{Fluck2018_NEON_Beetle}.} After fine-tuning, \OursA demonstrates segmentation results better than SegGPT~\cite{wang2023seggpt} and comparable with Mask2Former~\cite{cheng2022masked}. \OursA is applied to SAM 2~\cite{ravi2024sam} in this experiment.} 
        \label{fig-s4: beetle qualitative}
    \end{figure}

\section{Qualitative Results on Fine-Tuned Models}

\begin{figure*}[t]
        \centering
        \includegraphics[width=0.9\columnwidth]{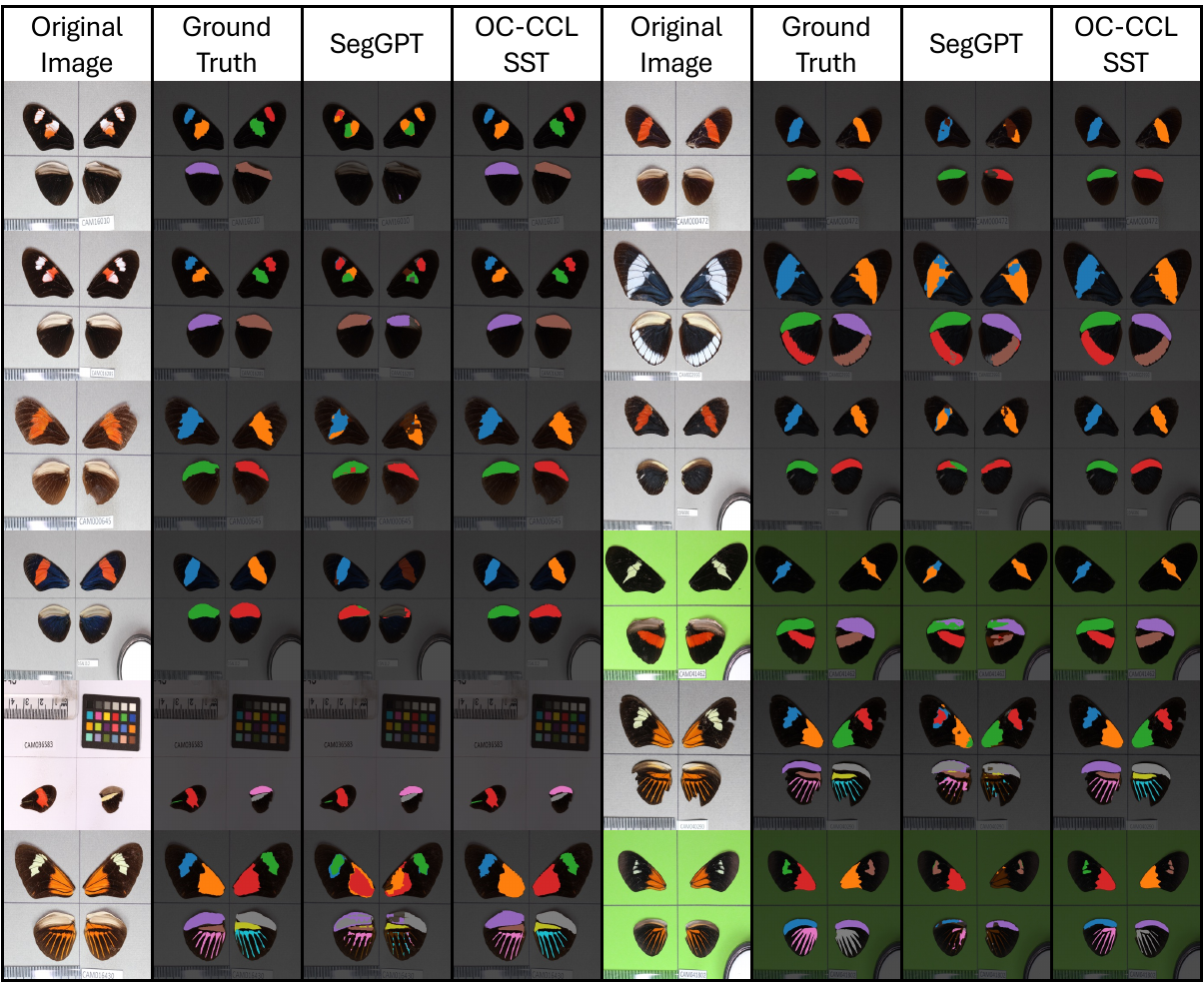}
        \caption{\small \textbf{Qualitative results on Butterfly~\cite{lawrence2024heliconius}.} \OursA and \OursL are applied to SAM 2~\cite{ravi2024sam}.}
        \label{fig-s3: butterfly qualitative}
    \end{figure*}

\begin{figure*}
        \centering
        \includegraphics[width=0.74\linewidth]{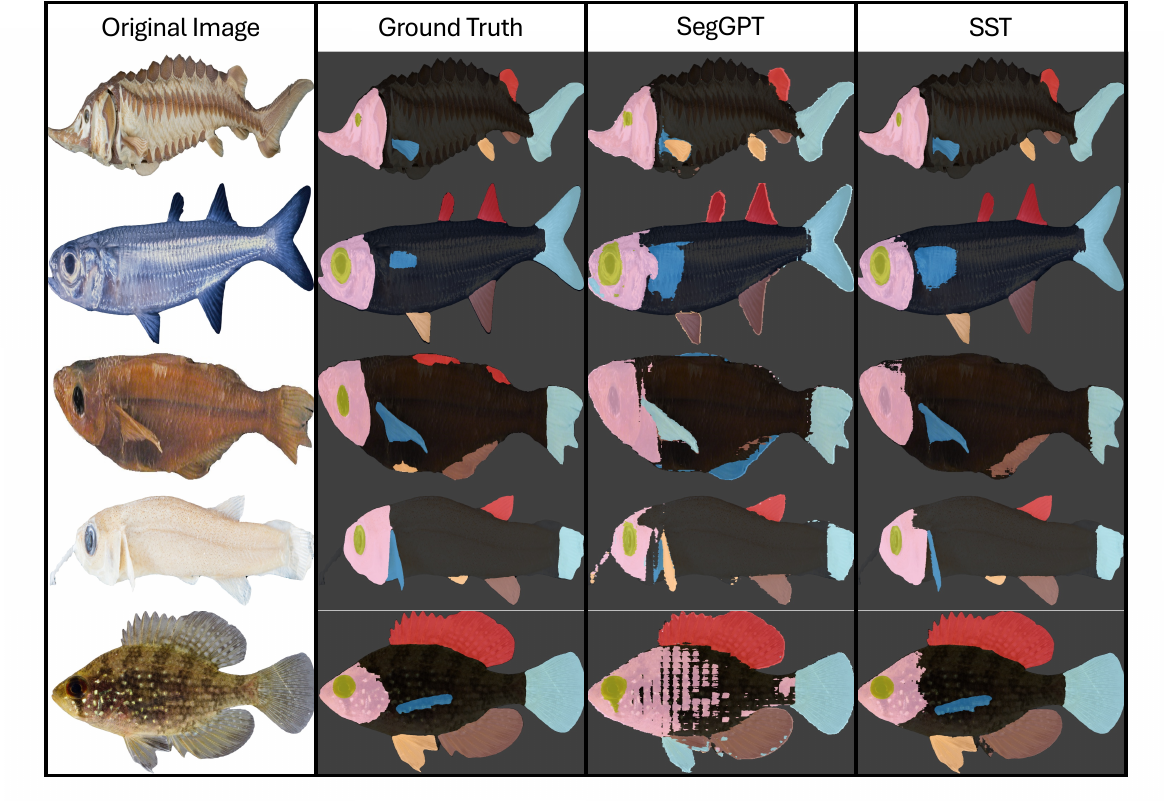}
        \caption{\small \textbf{Qualitative results on Fish~\cite{fishvistaData}.} \OursA and \OursL are applied to SAM 2~\cite{ravi2024sam}.}
        \label{fig-s2: fish qualitative}
    \end{figure*}

To compare the performance of different methods, we show more qualitative results on a more diverse set of butterflies, fish, and beetle species, see \cref{fig-s4: beetle qualitative}, \cref{fig-s3: butterfly qualitative}, and \cref{fig-s2: fish qualitative}. For each column, we keep the same setting as in the main paper. For both SegGPT~\cite{wang2023seggpt} and \OursA + \OursL, we select one random image from the training set as a reference and evaluate the segmentation quality on the target image. The quality is demonstrated in the third and fourth columns of the figures. We also show the segmentation quality of Mask2Former~\cite{cheng2022masked} in the last column of \cref{fig-s4: beetle qualitative} and \cref{fig-s2: fish qualitative}, which is trained on the entire available training dataset. We omit the Mask2Former column for the Butterfly~\cite{lawrence2024heliconius} dataset as there aren't enough data samples to train a full standard segmentation model for most of these subspecies.

\section{Additional Trait-Based Retrieval Results}

\begin{figure*}[t]
    \centering
    \includegraphics[width = 0.98\linewidth]{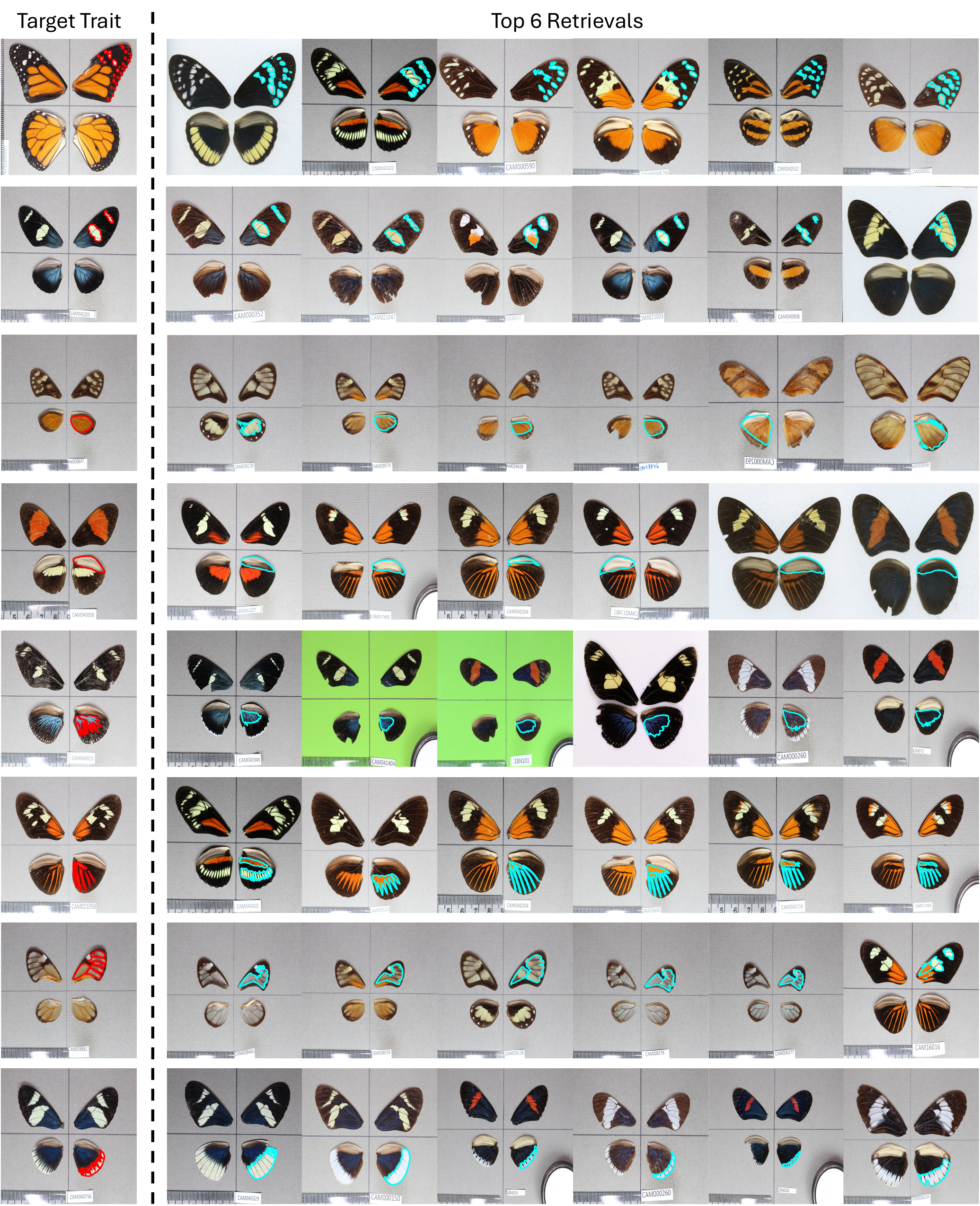}
    \caption{\small \textbf{Qualitative results for trait retrieval on Butterfly~\cite{lawrence2024heliconius}.} Target trait is outlined in \textbf{\textcolor{red}{red}}, the retrieved traits are outlined in \textbf{\textcolor{cyan}{cyan}}.}
    \label{fig-s5: supp trait retrieval}
\end{figure*}

We originally demonstrated \OursA's ability to do trait-based retrieval in Sections 3.4 and 4.5 in the main paper. Here, we show more trait-based retrieval results using \OursA with a diverse range of subspecies. Given a target trait on any subspecies, as outlined in red in the left-most column of \cref{fig-s5: supp trait retrieval}, \OursA can reliably retrieve subspecies that share similar traits, as outlined in cyan in \cref{fig-s5: supp trait retrieval}.

\end{document}